\documentclass[a4paper,man,natbib,floatsintext]{apa6}
\usepackage{apacite}
\usepackage{subfig}
\usepackage[english]{babel}
\usepackage[utf8x]{inputenc}
\usepackage{amsmath}
\usepackage{graphicx}
\usepackage[colorinlistoftodos]{todonotes}
\usepackage[left]{lineno}
\usepackage{diagbox}
\usepackage{indentfirst}
\usepackage{adjustbox}
\usepackage{xcolor}
\usepackage{multirow}
\usepackage{tabularx}

\usepackage{setspace}

\begin{document}

\raggedbottom

 \captionsetup[table]{
  labelsep = newline,
  textfont = sc, 
  name = TABLE, 
  justification=raggedleft,
  singlelinecheck=off,
  labelsep=colon,
  skip = \medskipamount}

\begin{titlepage}

\begin{center}

\large Predicting Driver Fatigue in Automated Driving with Explainability\\ 

\normalsize

\vspace{25pt}
Feng Zhou\\
Department of Industrial and Manufacturing Systems Engineering, \\The University of Michigan, Dearborn, MI, USA\\
email: fezhou@umich.edu\\
\vspace{15pt}
Areen Alsaid \\
Department of Industrial and Systems Engineering, \\The University of Wisconsin, Madison, WI, USA\\
email: alsaid@wisc.edu\\
\vspace{15pt}
Mike Blommer, Reates Curry, Radhakrishnan Swaminathan, Dev Kochhar, Walter Talamonti, and Louis Tijerina \\
Ford Motor Company Research and Advanced Engineering, Dearborn, MI, USA\\
email: \{mblommer, rcurry4, sradhak1, dkochhar, wtalamo1, ltijeri1\}@ford.com\\
\vspace{15pt}

\end{center}
\begin{flushleft}
\vspace{30pt}
\textbf{Manuscript type:} \textit{Research Article}\\
\textbf{Running head:} \textit{Fatigue Prediction with Explainability}\\
\textbf{Word count:} 5117 \\
\textbf{Acknowledgment}: This work was supported by Ford Summer Sabbatical Program \\
\textbf{Corresponding author:} 
Feng Zhou, 4901 Evergreen Road, Dearborn, MI 48128, Email: fezhou@umich.edu

\end{flushleft}

\end{titlepage}
\shorttitle{}

\section{Abstract}
Research indicates that monotonous automated driving increases the incidence of fatigued driving. Although many prediction models based on advanced machine learning techniques were proposed to monitor driver fatigue, especially in manual driving, little is known about how these black-box machine learning models work. In this paper, we proposed a combination of eXtreme Gradient Boosting (XGBoost) and SHAP (SHapley Additive exPlanations) to predict driver fatigue with explanations due to their efficiency and accuracy. First, in order to obtain the ground truth of driver fatigue, PERCLOS (percentage of eyelid closure over the pupil over time) between 0 and 100 was used as the response variable. Second, we built a driver fatigue regression model using both physiological and behavioral measures with XGBoost and it outperformed other selected machine learning models with 3.847 root-mean-squared error (RMSE), 1.768 mean absolute error (MAE) and 0.996 adjusted $R^2$.  Third, we employed SHAP to identify the most important predictor variables and uncovered the black-box XGBoost model by showing the main effects of most important predictor variables globally and explaining individual predictions locally. Such an explainable driver fatigue prediction model offered insights into how to intervene in automated driving when necessary, such as during the takeover transition period from automated driving to manual driving.

\textbf{Keywords: Driver fatigue prediction, explainability, automated driving, physiological measures} 

\newpage

%

\section{Introduction}
A driver may become fatigued or drowsy because of sleep deprivation, boredom, or monotony, time-on-driving tasks, medication side-effects, or a combination of such factors. Research indicates that the incidence of driver fatigue can be increased by monotonous automated driving \citep{vogelpohl2019asleep}. This can be dangerous in SAE Level 2 - Level 4 \citep{sae2018taxonomy} automated vehicles after the driver is out of the control loop for prolonged periods \citep{hadi2020influence}. Depending on the automation level of the vehicle, drivers need a high level of situation awareness in SAE Level 2 (partial automation) automated vehicles and good capabilities to respond to emerging hazards for takeover requests in SAE Level 3 (conditional automation) and Level 4 (high automation) automated vehicles \citep{collet2019associating}. For example, takeover requests will be issued in conditional automated driving, when the vehicle hits the operational limit, such as adverse weather conditions and construction zones \citep{ayoub2019manual,du2020examining,du2020predicting}, which require the driver to safely take over control from automated driving. Therefore, it is important to make sure that the driver is available and ready in certain situations in automated driving \citep{zhou2019takeover,du2020psychophysiological,du2020predicting1}. 


Although driver fatigue has been widely studied in manual driving (e.g., see \citep{dong2010driver,sikander2018driver}), the probe into the fatigue prediction in automated driving seems limited. In automated driving, many researchers instead focus on the influence of performing non-driving related tasks (e.g., cognitive workload, engagement, and distraction) on takeover performance \citep{du2020examining,clark2017performance}. On the other hand, if drivers are not involved in non-driving related tasks, they would quickly show signs of fatigue \citep{vogelpohl2019asleep}, which could potentially influence their takeover performance, too. For example, \citet{gonccalves2016drowsiness} found that participants felt subjectively fatigued even after as short as 15 minutes of a monitoring task in automated driving and \citet{feldhutter2017duration} identified fatigue indicators among 31 participants in a 20-minute automated driving scenario using eye-tracking data. Furthermore, in conditional automated driving, \citet{hadi2020influence} found that the higher the degree of fatigue was, the worse the takeover performance. Hence, it is critical to detect and predict driver fatigue in monotonous automated driving between SAE Level 2 and Level 4. 

Another phenomenon witnessed is that increasingly more researchers applied advanced machine learning models in driver fatigue detection and prediction in order to improve the performance of the models (see \citep{sikander2018driver}) due to their great successes in learning hidden patterns and making predictions of unobserved data, such as deep learning models based on convolutional neural networks (CNNs) and long short-term memory (LSTM). For example, \citet{dwivedi2014drowsy} used CNNs to explicitly capture various latent facial features to detect driver drowsiness. \citet{nagabushanam2019eeg} proposed a two-layer LSTM and four-layer improved neural network deep learning algorithm for driver fatigue prediction and their method outperformed other machine learning models. 

However, the trust and acceptance of such models can be compromised without revealing the domain knowledge, i.e., explainability or interpretations, contained in the data \citep{doshi2017towards}. 
Unlike other domains, the importance of explainable machine learning models in decision making with high risks is even greater, such as medicine \citep{lundberg2018explainable} and transportation \citep{zhou2020driver}. This is also advocated by \citet{mannering2020big} in safety analysis to consider both predictability and causality using advanced machine learning model. Furthermore, the domain knowledge captured by the machine learning models can be further used as guidelines to address the issues at hand. For example, \citet{caruana2015intelligible} built a generalized additive model with pairwise interactions to predict pneumonia
risks and found that those with asthma were less likely to die from pneumonia, which was counter-intuitive. However, by examining the data and the model, the researchers found that those with asthma were intensively cared, which was effective at reducing the likelihood of dying from pneumonia compared to the general population. Such knowledge explained the model behavior. Thus, similar knowledge can be potentially identified and used in driver fatigue prediction using explainable models in manual and automated driving to help provide effective intervention measures. 

Towards this end, we proposed an explainable machine learning model to predict driver fatigue using XGBoost (eXtreme Gradient Boosting) \citep{chen2016xgboost} and SHAP (SHapley Additive exPlanations) \citep{lundberg2018explainable,lundberg2020local} in automated driving. First, XGBoost is a highly effective and efficient algorithm based on tree boosting and it is one of the most successful machine learning algorithms in various areas, including driver fatigue prediction \citep{kumar2017predictive}. In order to understand the hidden patterns captured by the XGBoost model, SHAP \citep{lundberg2018explainable,lundberg2020local} was used to explain the XGBoost model by examining the main effects of the most important measures globally and explaining individual prediction instances locally. 
SHAP uses the Shapley value from cooperative game theory \citep{shapley1953contributions} to calculate individual contributions of the features in the prediction model and satisfies many desirable properties in explaining machine learning models, including local accuracy, missingness, and consistency \citep{lundberg2020local}. However, it is challenging to compute the exact Shapley values for features of machine learning models, especially deep learning models. \citet{lundberg2018consistent} proposed the SHAP algorithm to reduce the complexity of calculating Shapley value in algorithms based on tree ensembles from $O(TL2^M)$ to $O(TLD^2)$, where $T$ is the number of trees, $L$ is the largest number of leaves in the trees, $M$ is the number of the features, and $D$ is the maximum depth of the trees. Hence, XGBoost and SHAP were used in this paper to predict driver fatigue and uncover the hidden patterns in the machine learning model.

\section{RELATED WORK}

\subsection{Driver Fatigue Detection and Prediction}

\subsubsection{Manual Driving}
Driver fatigue has been studied widely in manual driving and previous studies examined driver fatigue from two main types of measures, including driving behavioral measures and physiological measures. Driving behavior measures mainly include steering motion and lane deviation \citep{koesdwiady2016recent}. For example, \citet{feng2009board} found that driver fatigue was negatively correlated with steering micro-corrections. \citet{sayed2001unobtrusive} proposed a fatigue prediction model with drivers’ steering angles based on an artificial neural network that classified fatigued drivers and non-fatigued driver with 88\% and 90\% accuracy among 12 drivers. Using a sleep deprivation study ($n$ = 12), \citet{krajewski2009steering} extracted features from slow drifting and fast corrective counter steering to predict driver fatigue and their best prediction accuracy was 86.1\% in terms of classifying slight fatigue from strong fatigue. \citet{li2017automatic} detected driver fatigue ($n$ = 10) by calculating approximate entropy features of steering wheel angles and yaw angles within a short sliding window with 88.02\% accuracy. \citet{mcdonald2014steering} applied a random forest steering algorithm to detect drowsiness indicated by lane departure among 72 participants and it performed better than other algorithms (e.g., neural networks, SVMs, boosted trees). 
Though driving behavioral measures are easier to collect, it is still challenging to obtain high prediction accuracy \citep{mcdonald2014steering}.

Many studies investigated driver fatigue using physiological measures, which have proven to be highly correlated with driver fatigue \citep{dong2010driver,sikander2018driver}. First, many researchers used head- and eye-related physiological data to detect fatigue \citep{ji2004real,watta2007nonparametric}. For instance, \citet{khan2008real} extracted features from the driver’ face and eyes to detect driver fatigue (indicated by eye closure) with a normalized cross-correlation function, which had 90\% accuracy. PERCLOS (percentage of eyelid closure over the pupil over time) and the average eye closure speed were used to detect driver fatigue using neural networks \citep{chang2014driver}. The system was able to detect fatigue with a success rate of 97.8\% among 4 participants. However, it dropped to 84.8\% when the participants wore glasses, and the reliability was susceptible to lighting, motion, and occlusion (e.g., sunglasses). Second, other popular methods used measures derived from EEG (electroencephalogram), EOG (electrooculography), ECG (electrocardiography), and EMG (electromyography) for fatigue detection. For example, \citet{jung2014driver} examined heart rate variability to monitor driver fatigue. \citet{zhang2013automated} extracted entropy and complexity measures from EEG, EMG, and EOG data of 20 subjects. \citet{lee2012driver} combined both photoplethysmography (PPG) signals and facial features to detect driver fatigue ($n$ = 10) and the model had a true and false detection rate of 96\% and 8\%, respectively, using a dynamic Bayesian network.

\subsubsection{Automated Driving}
Although these previous research endeavors provided insights into the progression of driver fatigue in manual driving, limited research is conducted in detecting and predicting driver fatigue in automated driving. \citet{gonccalves2016drowsiness} found that participants were easily fatigued due to underload in automated driving for 15 minutes of a monitoring task. Similarly, \citet{feldhutter2017duration} found fatigue signs due to underload in 31 participants using eye-tracking data for as short as 20-minute automated driving. Moreover, \citet{korber2015vigilance} found that participants ($n$ = 20) experienced substantial passive fatigue due to monotony after 42 minutes of automated driving using eye-related data. \citet{hadi2020influence} demonstrated that drivers' ($n$ = 12) takeover performance was significantly worse in various scenarios for fatigued driver in conditional automated driving. \citet{vogelpohl2019asleep} ($n$ = 60) indicated that compared to sleep-deprived drivers in manual driving, drivers in automated driving exhibited facial indicators of fatigue 5 to 25 minutes earlier and their takeover performance was significantly jeopardized. Therefore, fatigued drivers could be one of the safety issues in takeover transition periods where a high level of situation awareness is needed. These studies indicate the necessity for driver fatigue detection and prediction in SAE Level 2 - Level 4 automated driving. 

\subsection{Explainable Machine Learning Models}
To detect and predict driver fatigue, it is extremely important to develop accurate machine learning models in both manual driving and automated driving. For example, CNN was used to extract spatial facial features in detecting driver fatigue \citep{dwivedi2014drowsy} and LSTM was used to model temporal relations of physiological measures to detect driver fatigue
\citep{nagabushanam2019eeg}. Nevertheless, it can be \emph{difficult} to trust and accept such black-box models without revealing its domain knowledge captured by the models, especially with the risks associated with the decisions based on the models are high. Therefore, the choice between simple, easier to interpret models and complex, black-box models is one of the important factors to consider in deploying such models. Usually there are two types of explainable models, i.e., model-based and post-hoc explainability \citep{murdoch2019definitions}. Typical examples of the model-based explainability include linear regression models, logistic regression models, and single decision tree models. However, their performance is usually inferior compared to complex black-box models. Post-hoc explainability is then used to explain the behaviors and working mechanisms of black-box models approximately, such as SHAP \citep{lundberg2018explainable,lundberg2020local} and LIME (Local Interpretable Model-agnostic Explanations) \citep{ribeiro2016should}. For example, LIME was used to explain neural network models in credit scoring applications \citep{munkhdalai2019advanced} and SHAP was used to explain ensemble machine learning models to identify risk factors during general anesthesia \citep{lundberg2018explainable}. Such explanation not only identified the key variables in modeling, but also increased trust in real applications \citep{AYOUB2021102}. Compared to LIME, SHAP was better in explaining machine learning models in terms of local accuracy and consistency \citep{lundberg2020local}.

\section{EXPERIMENT DESIGN}
\subsection{Participants}
We excluded participants who had caffeine consumption, sleep disorders or other factors that might have impacted driver fatigue. Finally, twenty participants were recruited in this study (14 males and 6 females between 20 and 70 years old). In order to elicit fatigue, unlike the common sleep deprivation method, we made use of the nature of underload and monotony in automated driving to elicit passive fatigue in the experiment in the afternoons. According to previous studies \citep{gonccalves2016drowsiness,feldhutter2017duration,korber2015vigilance}, participants were expected to show passive fatigue signs as soon as in 15 minutes without doing any secondary tasks and such fatigue was more prevalent in automated driving.

\subsection{Apparatus}
\begin{figure} [H]
\centering
\includegraphics[width=0.8\columnwidth]{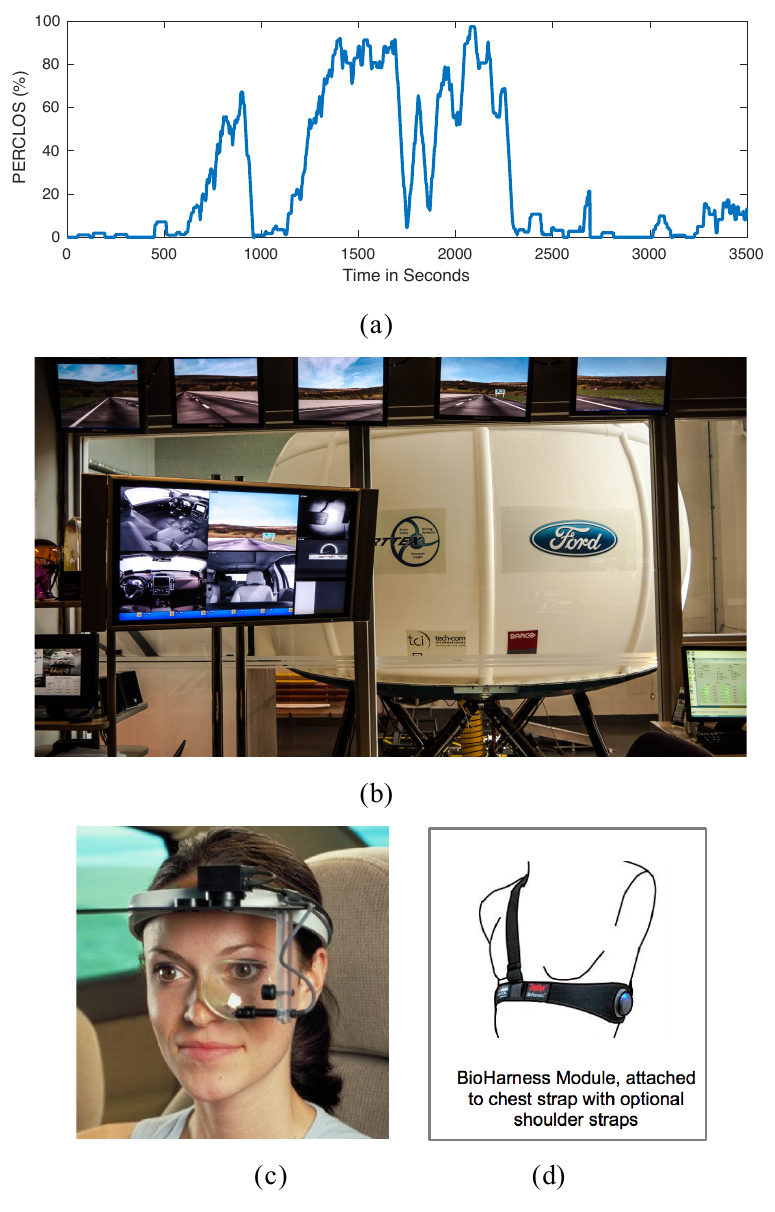}
\caption{(a) Example PERCLOS time series data for a participant. (b) The VIRTTEX driving simulator and driving scenarios with different views. (c) The configuration of ISCAN Eye Tracker for PERCLOS measure collection. (d) BioHarness model for physiological data collection.}
\label{fig:Exp}
\end{figure}

The study took place in the VIRTTEX (VIRtual Test Track EXperiment) driving simulator at Ford (see Figure \ref{fig:Exp}b), a large six degree-of-freedom motion base simulator that uses a hydraulically powered Stewart platform to reproduce vehicle motion. The visual environment consists of a 240° front field-of-view and a 120° rear field-of-view. Drivers were seated in a Ford Edge cab with 3D simulated sound to provide realistic interior and exterior environment sounds, as well as a steering control loader for accurate road feedback and tire forces to the driver. The simulator was configured with an SAE Level 3 automated driving system and auditory-visual displays to indicate automated system status throughout the drive. The driver wore ISCAN© eye-tracking goggles (ISCAN, Inc., MA, USA) outfitted with an eye camera, a dichroic mirror, and a scene camera to track percent pupil occlusion in real time in order to calculate the PERCLOS measure (see Figure \ref{fig:Exp}c). Before entering the simulator, the participant was outfitted with a BioHarness 3.0 Wireless Heart Rate Physiological Monitor (Zephyr Technology, MD, USA) with Bluetooth to capture physiological measures (see Figure \ref{fig:Exp}d). 

\subsection{Simulator Scenario}
The simulated drive was on a 4-lane undivided rural roadway with light traffic flowing with and opposing the drivers. Drivers were instructed to initially stay in the right lane and follow a lead vehicle that was varying its speed between 50 – 70 mph (80 – 115 kph). The driver then engaged the automated driving system using the steering wheel controls. A key feature of this scenario was that, after the automation was engaged for the data collection portion of the study, there were no secondary tasks at all or no interaction with the research staff until the end of the session in order to elicit passive driver fatigue in automated driving. There were also no hazardous events in the automated driving session, which lasted about 60 minutes. With essentially no disruptions, some drivers entered the fatigued state quickly \citep{gonccalves2016drowsiness,feldhutter2017duration,korber2015vigilance}.    

\subsection{Predictor Variables and Response Variables}
The response variable to be predicted was PERCLOS obtained from the ISCAN eye tracker sampled at 60Hz. PERCLOS was operationally defined as the average percent of the time the eyelids occluding the pupil (larger than 80\%) using a 1-minute moving window at any point in the data collection session during the experiment \citep{zhou2020driver}. A typical example of a fatigued participant was shown in Figure \ref{fig:Exp}a. 
 
The reason that we used PERCLOS as the ground truth of our prediction model was that it was a reliable indicator of driver fatigue \citep{zhou2020driver}, but it was intrusive to measure in real applications (Figure \ref{fig:Exp} (c)). Therefore, we collected 11 less intrusive measures as predictor variables as shown in Table \ref{tab:predictor} to predict driver fatigue indicated by PERCLOS. A low pass filter was used to remove baseline wander noises in ECG data \citep{kher2019signal}. Breathing wave signals were filtered using a moving average filter to remove noise. Steering wheel angles (swa), torque applied on steering wheel (intertq), and posture data were also filtered by a low pass filter. Other measures, such as hr\_{avg60}, were then calculated based on the filtered signals. 

\begin{table} [H]
\caption{Predictor variables used in modeling driver fatigue}
\label{tab:predictor}
\centering
\begin{tabularx}{\columnwidth} {llX}
    \hline
    Features & Unit &    Explanation \\
    \hline
    heart\_{rate}\_{variability}   &   millisecond (ms) & {Standard deviation of inter-beat interval}\\
    hr\_{avg60}   &   beats/minute & {Average heart rate with a 60s sliding window}\\
    br\_{avg60}   &   breaths/minute & {Average breathing rate with  a 60s sliding window}\\
    br\_{std60}   &   breaths/minute & {Standard deviation of breathing rate with a 60s sliding window}\\
    hr\_{std60}   &   beats/minute & {Standard deviation of heart rate with a 60s sliding window}\\
     heart rate   &   beats/minute & {Number of beats in one minute} \\
     breathing   &   bits & Breathing waveform (16Hz) \\
     ECG   &   mV & ECG waveform (250Hz)\\
     intertq   &   Nm & {Torque applied to steering wheel (200Hz)} \\
     swa   &   degree & {Steering wheel angle (200Hz)} \\
     posture   &   degree & {Degree from subject vertical (1Hz)} \\
    \hline 
\end{tabularx}
\end{table}

\subsection{Experimental Procedure}
Once participants had signed the informed consent form, they were given a brief study introduction via a PowerPoint® presentation. The study introduction highlighted the study objective to examine driver reactions to an automated driving system. Participants were then given an overview of the automated driving system’s capabilities and training on how to engage and disengage the system.  Prior to entering the VIRTTEX simulator, the participant donned the BioHarness 3.0 belt and Bluetooth connectivity was verified. Once seated in the cab, the participant put on the eye-tracking goggles. Up to 10 minutes was spent on calibrating the eye-tracker (typically less than 5 minutes for drivers without glasses). Once the training was complete, the participants transitioned right into the main drive. The driver was instructed to engage the automation, monitor the driving task, and not to do any secondary tasks or communicate with the experimenter. 

\section{Driver Fatigue Prediction and Explanation}
\subsection{XGBoost}
XGBoost is a highly efficient and effective machine learning model both for regression and classification \citep{chen2016xgboost}. In our driver fatigue prediction, we used a one-second time window to discretize all the predictor variables, $\mathbf{X} = \{\mathbf{x}_k\}$, and response variable, $Y = \{y_{k}\}$, of all the participants, where $k = 1, ..., n$. The training data set is indicated as $D = { \{\mathbf{x}_{k}, y_{k},  \mathbf{x}_{k}  \in  R^{m}, y_{k} \in R\}}$. In this research, $n$ is the total number of the samples and $m$ is the number of the features (i.e., predictor variables), and $n = 58846, m = 11$. Let $\hat{y}_{k}$ denote the predicted result of a tree-based ensemble model,  $\hat{y}_{k} =  \phi ( x_{k} ) =  \sum_{s=1}^S  f_{s}( x_{k} )$, where $S$ is number of the trees in the regression model, $f_{s}\big(x_{k}\big)$ is the $s$-th tree. For XGBoost, the objective function is regularized to prevent over-fitting as follows:
\begin{equation} \label{objective}
 L\big( \phi \big) =  \sum_k  l\big(y_{k},\hat{y}_{k}\big) +  \sum_s \Omega \big(f_s\big), 
\end{equation}
where $l$ is the loss function and in this research, we used root-mean-squared error (RMSE). The penalty term $\Omega$ has the following form:
\begin{equation} \label{regulizer}
 \Omega \big(f_s) =  \gamma T+ \frac{1}{2} \lambda  \| w \| ^{2}, 
\end{equation}
where $\gamma$ and $\lambda$ are the penalty parameters to control the number of leaves $T$ and the magnitude of leaf weights $w$. In the training process, XGBoost used an iterative process to minimize the objective function and for the $i$-th step when adding a tree, $f_{i}$, to the model as follows:
\begin{equation} \label{iterative}
 L^i\big( \phi \big) =  \sum_k  l\big(y_{k},\hat{y}_{k}^{(i-1)}+f_{i}(x_{k})\big) +  \Omega \big(f_i\big), 
\end{equation}
This formula was approximated with a 2nd order Taylor expansion by substituting the loss function with mean-squared error and after tree splitting from a given node, we have:
\begin{equation} \label{splitting}
\begin{aligned}
L_{split} =  \frac{1}{2} \big(\frac{ (\sum_{k \in K_{L}} g_{k})^2  }{(\sum_{k \in K_{L}} h_{k} + \lambda )}  + \frac{ (\sum_{k \in K_{R}} g_{k})^2 }{(\sum_{k \in K_{R}} h_{k} + \lambda )} - \frac{\sum_{k \in K} g_{k})^2 }{(\sum_{k \in K} h_{k} + \lambda )} \big)  -\gamma,
\end{aligned}
\end{equation}
where $K$ is a subset of observations for the given node and $K_{L},K_{R}$ are subsets of observations in the left and right trees, respectively. $g_{k}$ and $h_{k}$ are the first  and second order gradient statistics on the loss function and are defined as $g_{k} = \partial _{ \hat{y}^{(j-1)}}l\big(y_{k},\hat{y}_{k}^{(j-1)}\big), h_{k} =  \partial^2_{\hat{y}^{(j-1)}}l\big(y_{k},\hat{y}_{k}^{(j-1)}\big)$. For a given tree structure, the algorithm pushed $g_{k}$ and $h_{k}$ to the leaves they belong to, summed the statistics together, and used Eq. (4) to identify the optimal splitting, which was similar to the impurity measure in a decision tree, except that XGBoost also considered model complexity in the training process.

\subsection{SHAP}
SHAP uses Shapley values \citep{shapley1953contributions} based on coalitional game theory to calculate individual contributions of each feature, which is named as SHAP values. In this research, SHAP was used to explain the main effects in the XGBoost model and individual predictions.  According to \citep{lundberg2020local,lundberg2018explainable}, SHAP values are consistent and locally accurate individualized features that obey the missingness property. The definition of the SHAP value of a feature-value set is calculated for a model $f$ below:
\begin{equation} \label{main}
\varphi_i(v)=\sum_{S\subseteq N\setminus \{i\}} {\frac {|S|!(m-|S|-1)!}{m!}}(f_{x}(S\cup \{i\})-f_x(S)), 
\end{equation}
where $f_{x}(S) = E[f(x)|x_S]$ is the contribution of coalition $S$ in predicting driver fatigue, indicated by PERCLOS in this study, $S$ is a subset of the input features, $N$ is the set of all the input features, and $m = 11$ is the total number of features. The summation extends over all subsets $S$ of $N$ that does not contain feature $i$. However, it is challenging to estimate the value of $f_{x}(S)$ efficiently due to the exponential complexity in Eq. (5). \citet{lundberg2018explainable} proposed an algorithm to approximate the values of $E[f(x)|x_S]$ for tree-based models, such as XGBoost, in $O(TLD^2)$ time, where $T$ is the number of the trees, $L$ is the number of maximum leaves in any tree, and $D = logL$. 
SHAP uses the difference between individual fatigue prediction against the average fatigue prediction, which is fairly distributed among all the feature-value sets in the data (see Figure \ref{fig:individual}). Hence, it has a solid theory foundation in explaining our XGBoost model. 

\section{Results}
\subsection{Prediction Results by XGBoost}
First, we compared the performance of the XGBoost prediction model with other six regression models, including linear regression, linear SVM, quadratic SVM, Gaussian SVM, decision trees, random forest. The setting of the XGBoost was as follows: max depth = 10, learning rate  = 0.1, objective = reg:squarederror, number of estimators = 150, regularization parameter alpha = 1, subsample = 0.9, and colsample = 0.9. All the 11 predictor variables were included in all the models with 10-fold cross validation, except the last entry for XGBoost (best), where only 5 most important features were selected to obtain the best performance (see Figure \ref{fig:Performance}). 
We reported the results in predicting PERCLOS (0-100) with the following three performance metrics, including RMSE (the smaller the better), MAE (i.e., mean absolute error, the smaller the better), and adjusted $R^2$ (the closer to 1, the better) defined as follows:
\begin{equation} \label{rmse}
   RMSE = \sqrt{\frac{\sum_{k=1}^n \big(y_{k}-\hat{y}_{k}\big)^2}{n} } , 
\end{equation}
\begin{equation} \label{mae}
   MAE = \frac{\sum_{k=1}^n |y_{k}-\hat{y}_{k}|}{n} , 
\end{equation}
\begin{equation} \label{adjustedR_squared}
Adj.~ {R}^2 = 1-(1-R^2) \frac{n-1}{n-m-1},
\end{equation}
where $n$ is the total number of the samples, $\hat{y}_k$ is the predicted value of the ground truth, $y_k$, and $R^2 =\frac{SS_{regression}}{SS_{total}}=
\frac{\sum_{k=1}^n \big(\hat{y}_{k} - \overline{y}\big)^2}
{\sum_{k=1}^n \big(y_{k}- \overline{y} \big)^2}$, and $m$ is the total number of the predictor variables. The results are shown in Table \ref{tab:result1}. It can be seen that XGBoost outperformed other machine learning models. 
\begin{table}[h]
\caption{Comparisons of prediction results of different machine learning models}
\label{tab:result1}
\centering
\begin{tabular}{p{5cm}p{1.7cm}p{1.7cm}p{1.7cm}c}
\hline
Model & RMSE & MAE & Adj. $R^2$ \\ \hline 
Linear Regression & 26.429 & 20.189 & 0.250 \\
Linear SVM & 28.972 & 18.793 & 0.102 \\
Quadratic SVM & 25.995 & 16.537 & 0.269 \\
Gaussian SVM & 18.027 & 11.915 & 0.653 \\
Fine Tree & 6.753 & 2.516 & 0.951 \\
Random Forest & 6.700 & 3.910 & 0.950 \\
XGBoost (all) & 4.788 & 2.316 & 0.993 \\
XGBoost (best) & \textbf{3.847} & \textbf{1.768} & \textbf{0.996} \\
\hline
\end{tabular}
\\Note XGBoost (all) indicates all the 11 predictors were included in the model and XGBoost (best) indicate only five of the most important features were included (see Figure \ref{fig:Performance}). 
\end{table}

\subsection{SHAP Explanation}
\subsubsection{Feature Importance}

\begin{figure}[h]
\centering
\includegraphics[width=\columnwidth]{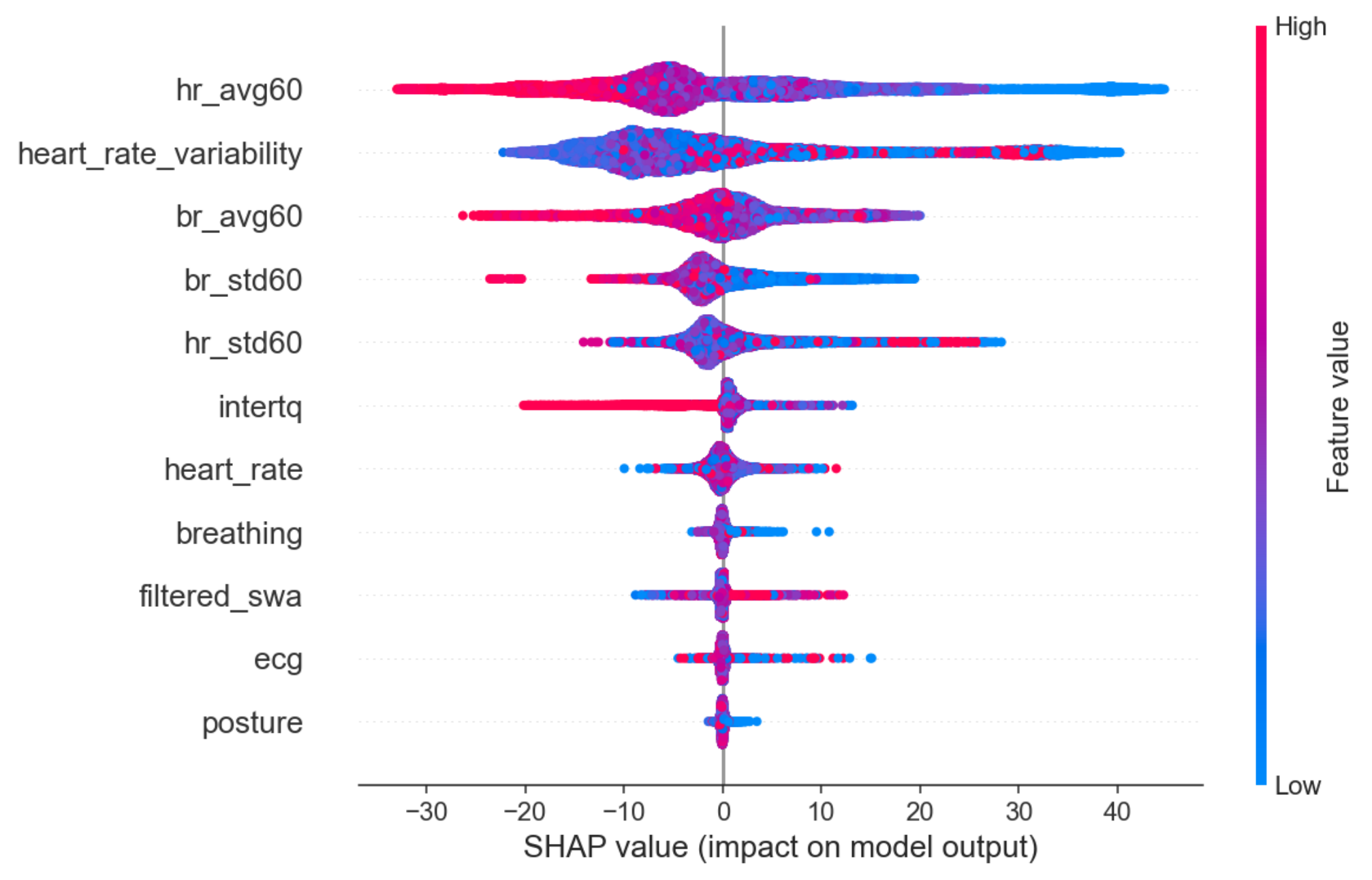}
\caption{Importance ranking of 11 features identified by SHAP summary plot. The higher the SHAP value of a feature, the higher the predicted PERCLOS.}
\label{fig:importance}
\end{figure}

During the 10-fold cross-validation process, we used the test data in each fold to calculate the SHAP values in order to improve its generalizability so that each sample was calculated its SHAP value for exactly once. We used the SHAP values in Eq. \ref{main} to identify the most important features as shown in Figure \ref{fig:importance}. The feature importance was sorted by their global impact  $\sum_{k=1}^n|\varphi_{k}^{m}|$ identified by SHAP plotted vertically as follows: hr\_avg60 is the most important, followed by heart\_rate\_variability, br\_avg60, br\_std60, hr\_std60, and so on. Every sample in the data was run through the model and each dot (i.e., $\varphi_{k}^{m}$) was created for each feature value and was plotted horizontally. The more important the feature is, the more impact on the model output. For example, hr\_avg60 had a range of SHAP value (i.e., PERCLOS) between -40 and 60. Note that the SHAP value was computed with regard to the base average output (see Figure \ref{fig:individual}) and hr\_avg60 could push some extreme output 40 lower than the average and push some other extreme output 60 higher than the average. It tended to show that the higher the value of hr\_avg60, the lower the predicted PERCLOS. 

\begin{figure} [h]
	\centering
	\subfloat[\label{fig:Performance1}]{\includegraphics[width=0.49\linewidth]{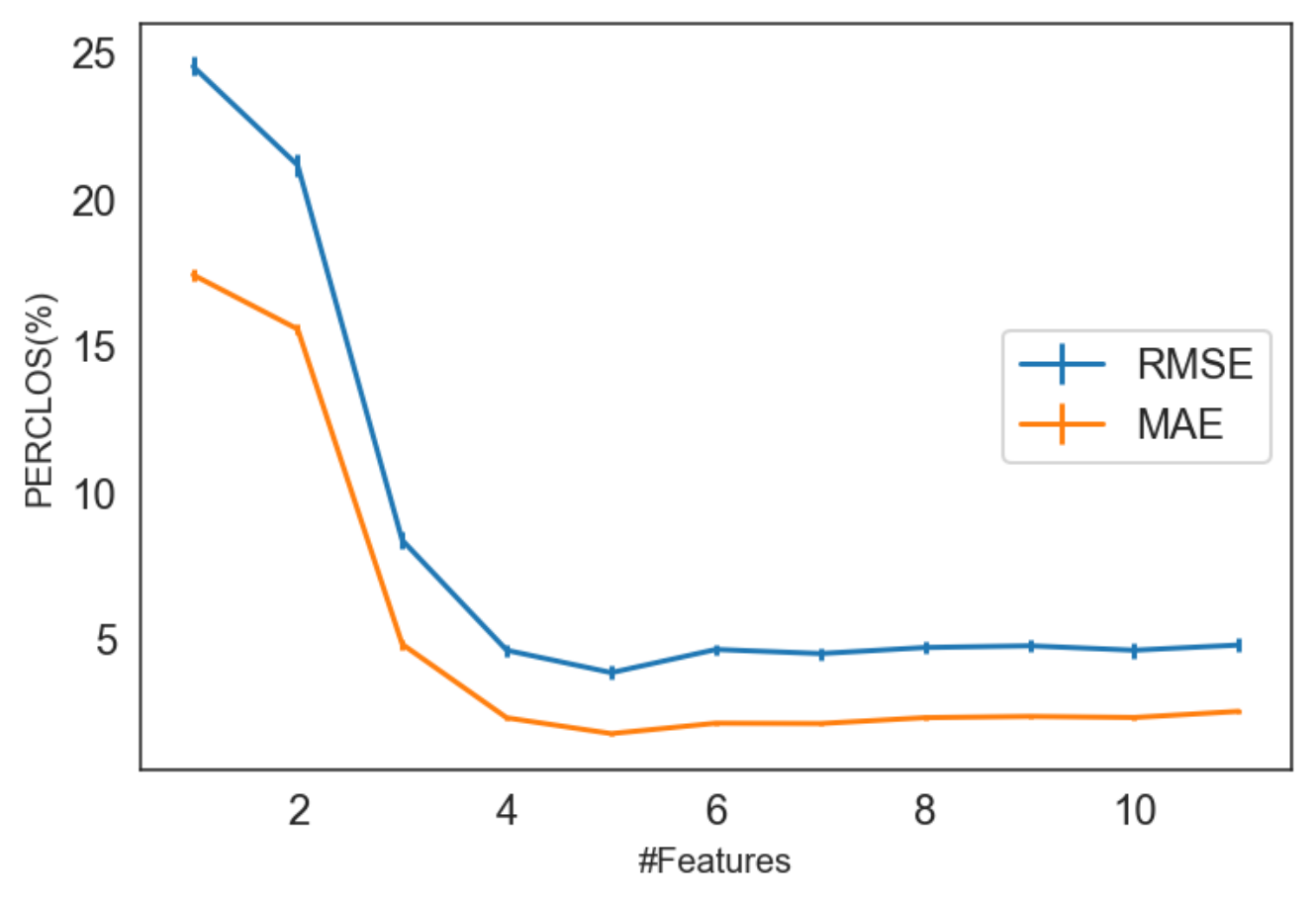}}
	\hspace{0pt}
	\subfloat[\label{fig:Performance2}]{\includegraphics[width=0.49\linewidth]{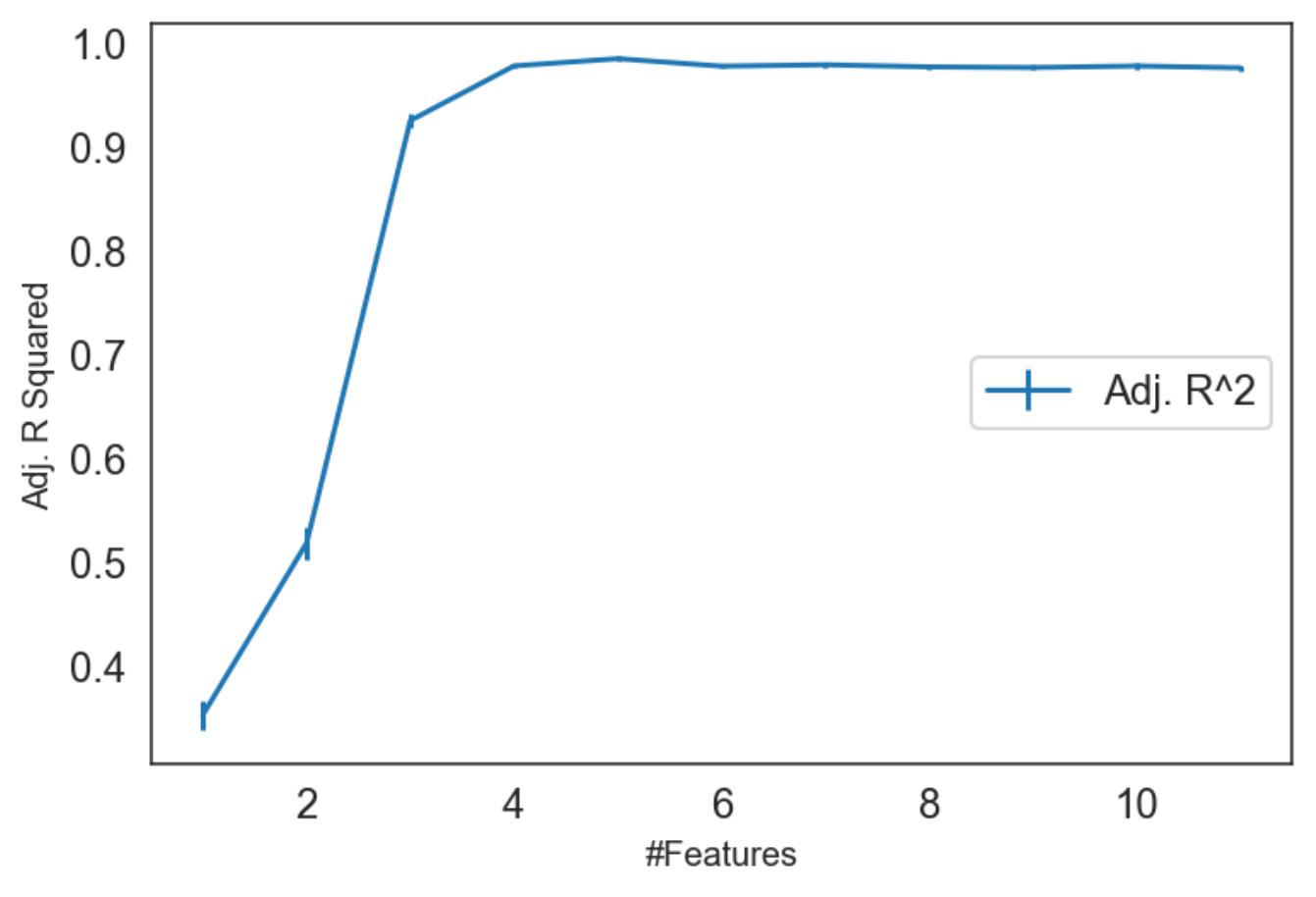}}
    \caption{How performance changes when the model added one feature at a time from the most important  one  to  the  least  important  one:  (a)  RMSE and MAE;  (b) Adjusted $R^2$. Note the error bar was the standard deviation obtained in the ten-fold cross validation process.}\label{fig:Performance}
\end{figure}

\subsubsection{Prediction Results with an Optimal Subset of Features}
We further added one feature at a time to the XGBoost model starting from the most important identified as shown in Figure \ref{fig:importance}. Figure \ref{fig:Performance} shows that the performance was increasing when more features were added until when there were 5 features (i.e., hr\_{avg60}, heart\_{rate}\_{variability}, br\_{avg60}, br\_{std60}, and hr\_{std60}) in the prediction model. The performance was better than that obtained by the model when 11 predictors were included, i.e., a subset of important features had the optimal performance (see Table \ref{tab:result1}).

\subsubsection{Main Effects}
\begin{figure} [H]
	\centering
	\subfloat[\label{fig:HR}]{\includegraphics[width=.42\linewidth]{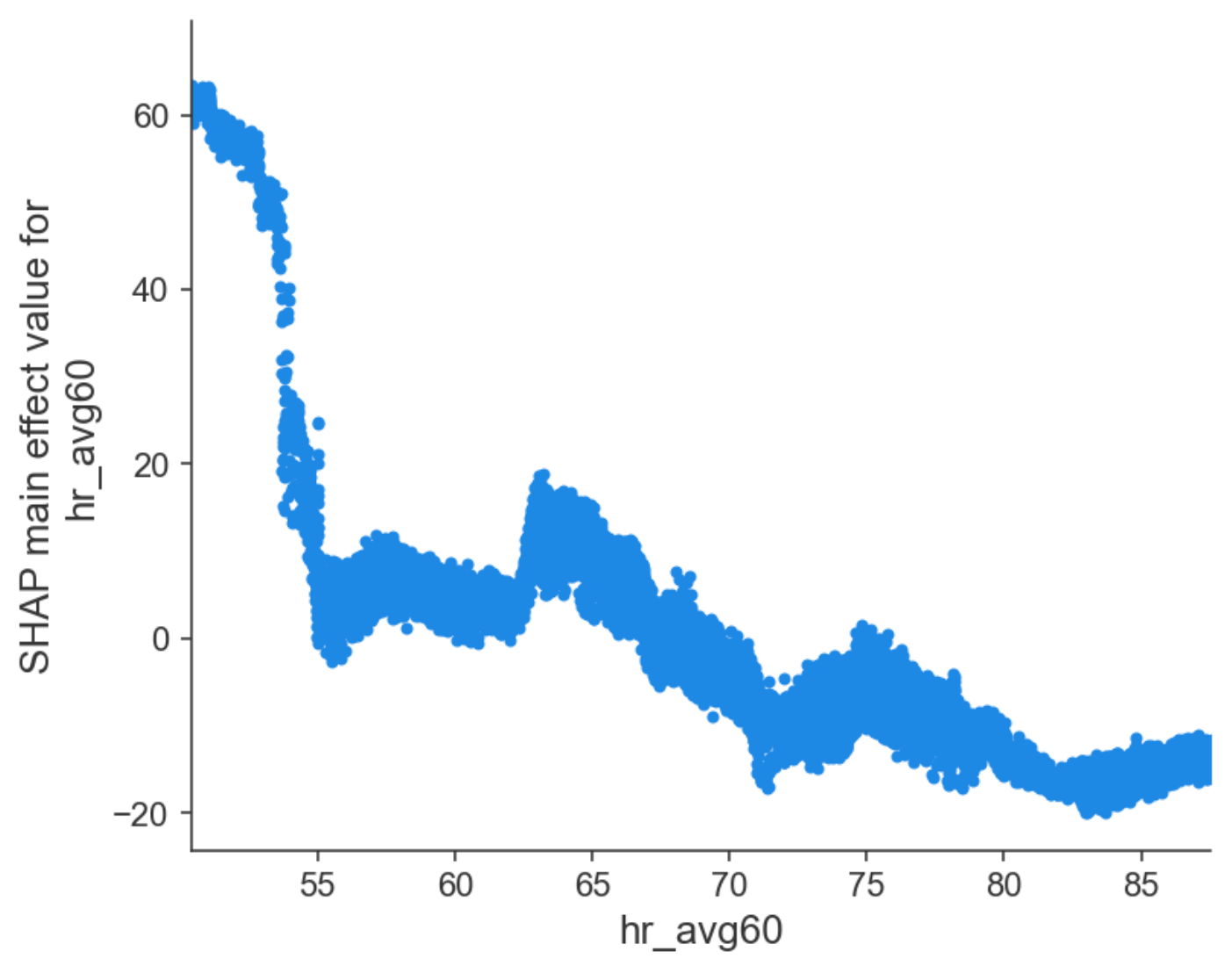}}
	\vspace{0pt}
	\subfloat[\label{fig:HRV}]{\includegraphics[width=.42\linewidth]{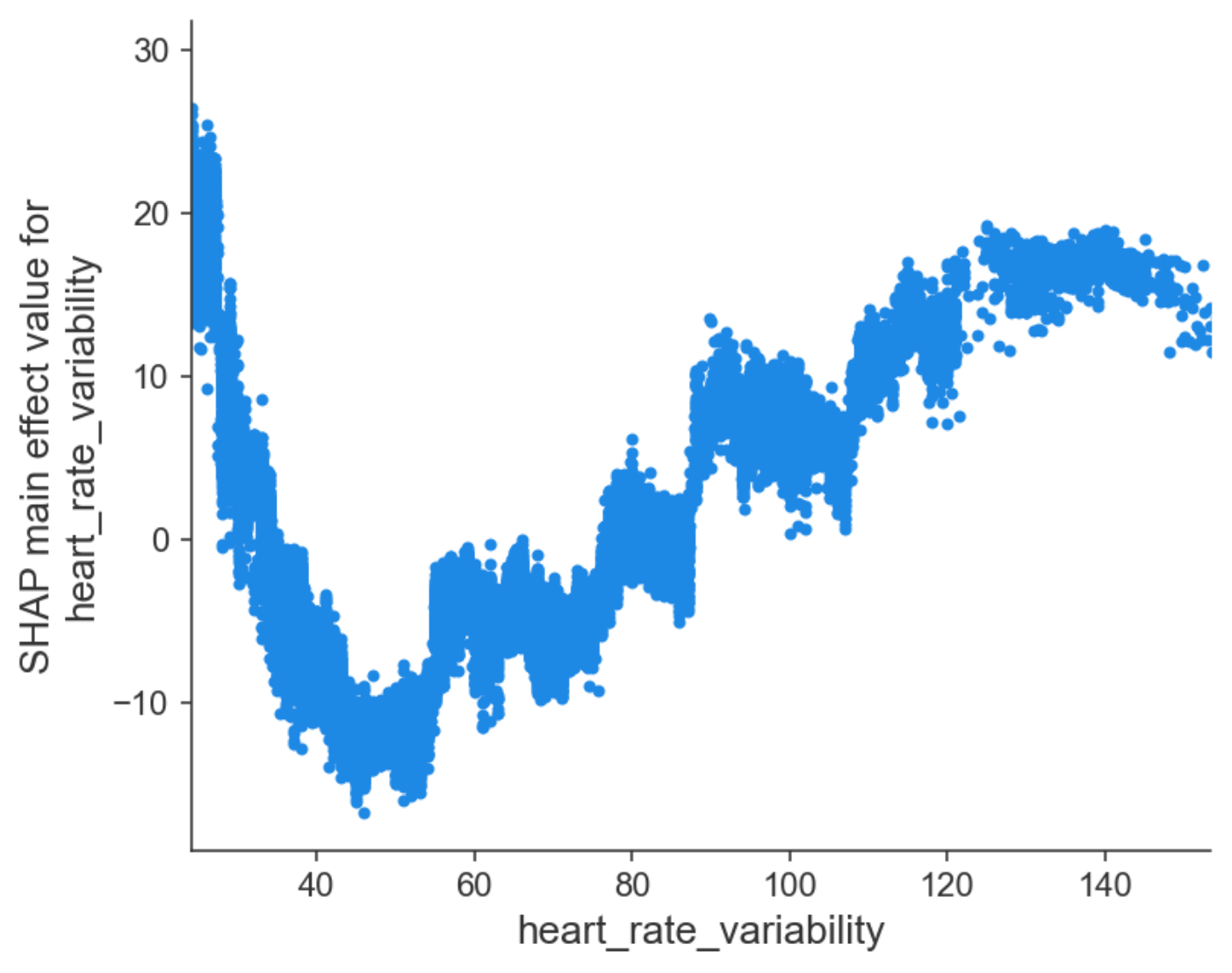}}
	\subfloat[\label{fig:BR}]{\includegraphics[width=.42\linewidth]{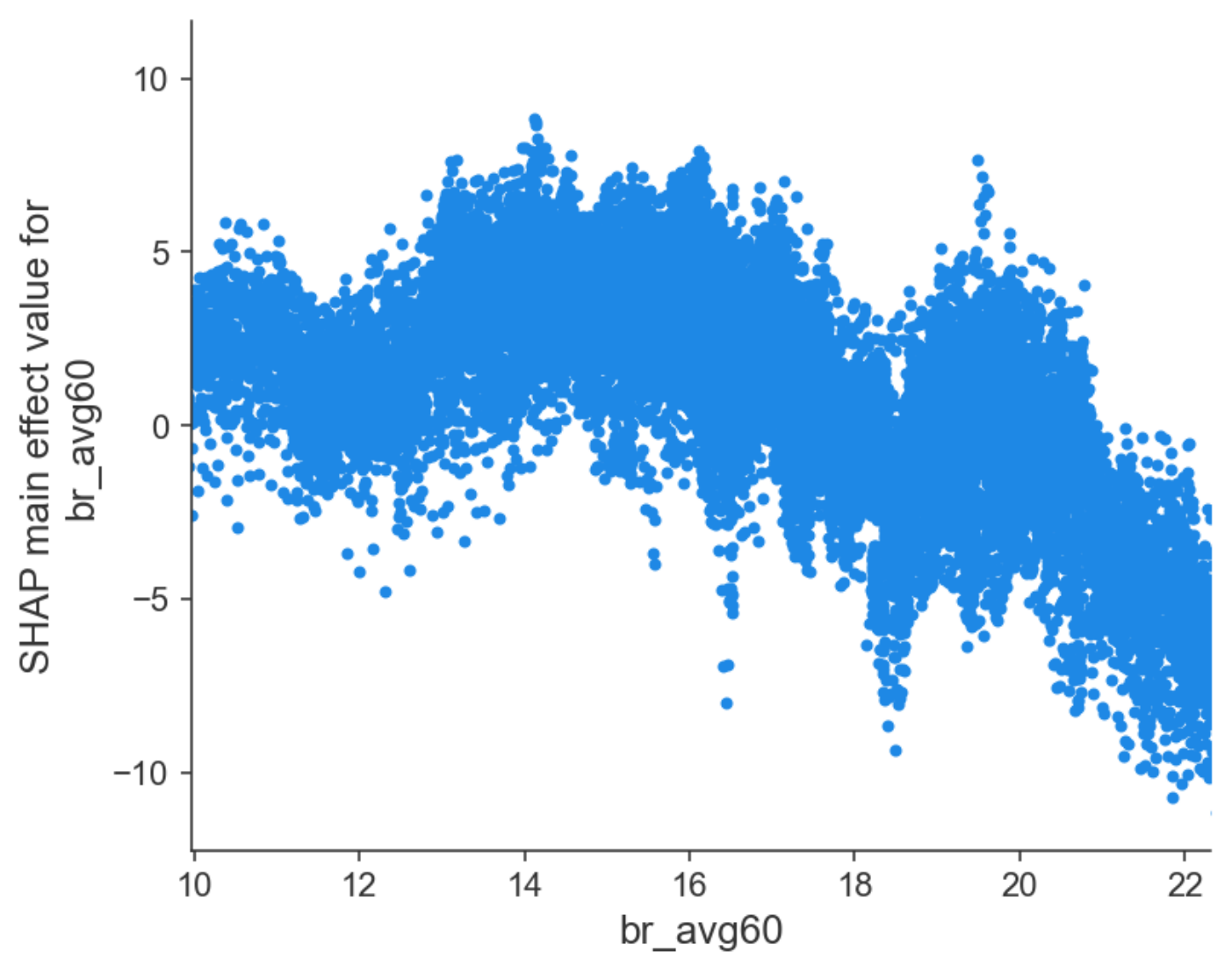}}
	\vspace{0pt}
	\subfloat[\label{fig:BRStd}]{\includegraphics[width=.42\linewidth]{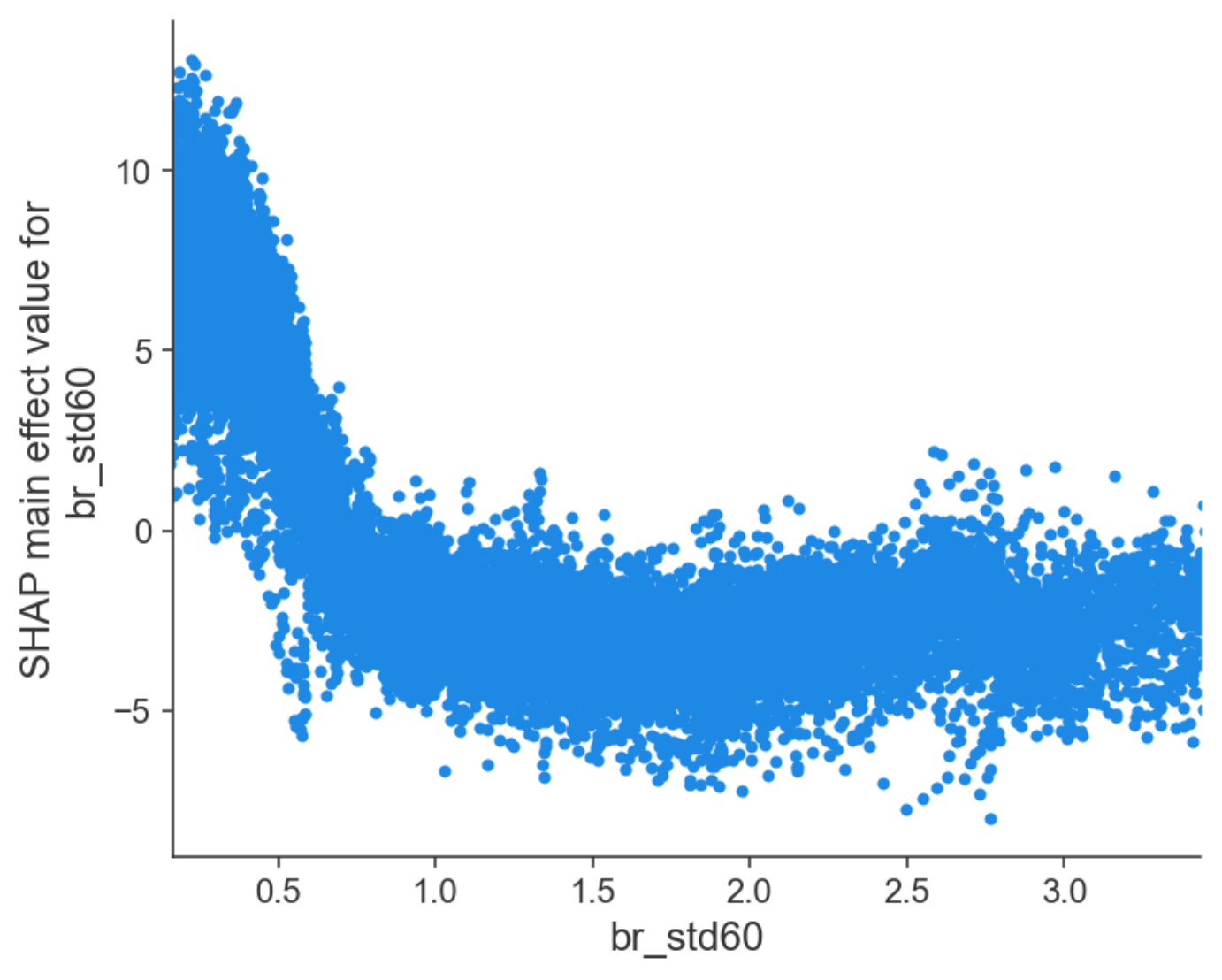}}
	\subfloat[\label{fig:HRStd}]{\includegraphics[width=.42\linewidth]{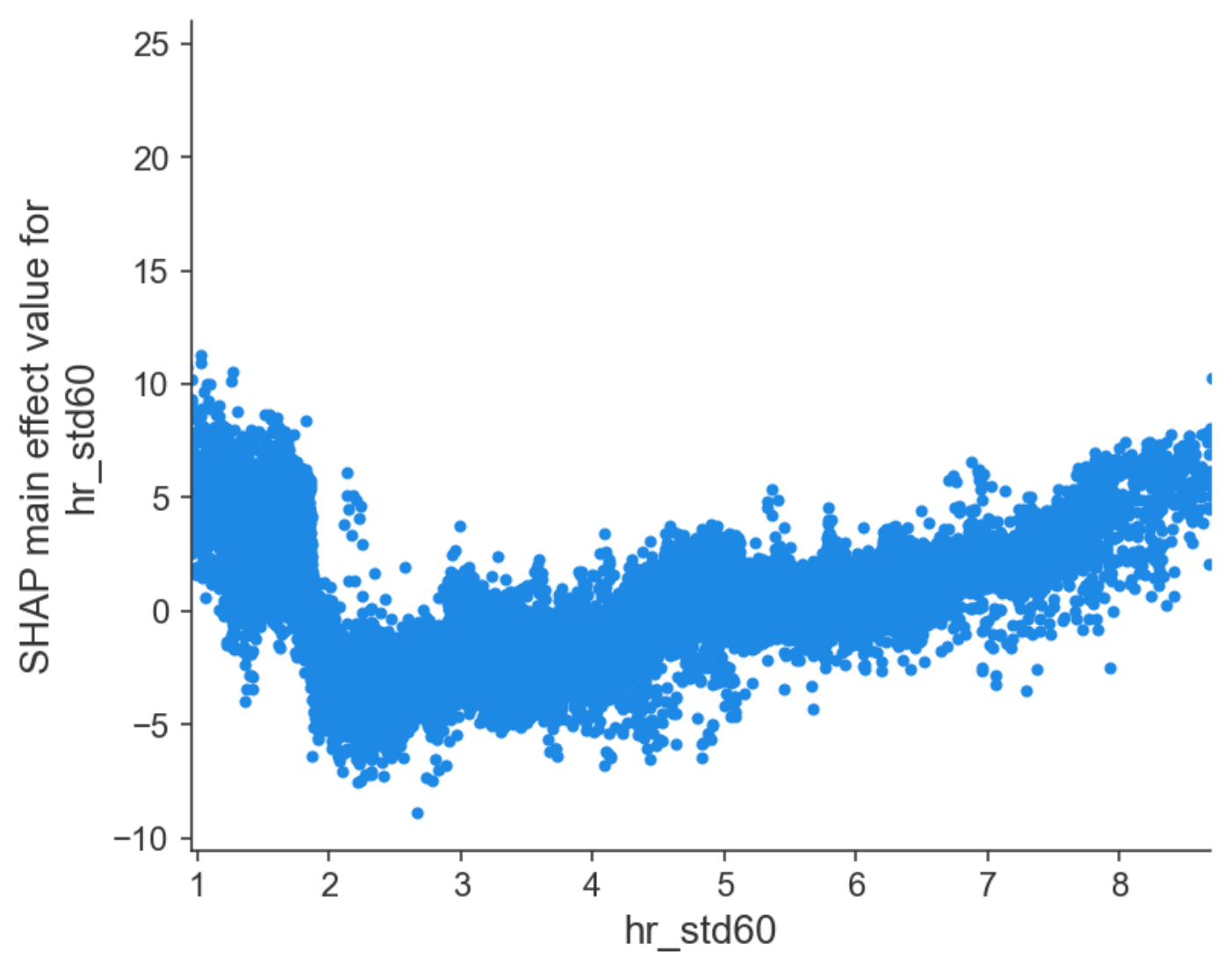}}
    \caption{Main effects of the most important features. (a) hr\_{avg60}; (b)  heart\_{rate}\_{variability}; (c) br\_{avg60}; (d) br\_{std60}; (e) hr\_{std60}. Note only data between the 2.5th percentile and the 97.5th percentile were included in the figures.}\label{fig:MainEffects}
\end{figure}

We also examined the main effects of the top five most important features when the model had the best performance in Figure \ref{fig:Performance}. Figure \ref{fig:MainEffects} shows the main effects. Consistent with Figure \ref{fig:importance}, the overall trend is that the larger the value of hr\_{avg60}, the smaller the SHAP values (i.e., predicted PERCLOS), but not in an exact linear fashion (see Figure \ref{fig:BR}). The slope is much larger in a narrow interval around 50 and 55 beats/min than others and between 55 and 63 beats/min, it is almost flat. When it is larger than 63 beats/min, the larger the value of hr\_{avg60}, the smaller the predicted PERCLOS. For heart\_{rate}\_{variability}, it has a V-shape relationship with the predicted PERCLOS (see Figure \ref{fig:HRV}). The predicted PERCLOS is decreasing when the value of heart\_{rate}\_{variability} is smaller than about 50 ms while the predicted PERCLOS is increasing when it is going up from 50 ms to 140ms. The overall trend for br\_{avg60} is that the larger the value of  br\_{avg60}, the smaller the predicted PERCLOS, and this trend was not obvious until the value of br\_{avg60} is larger than 15 breaths/min (see Figure \ref{fig:BR}). The overall trend for br\_{std60} is that the predicted PERCLOS is decreasing when the value of br\_{std60} is increasing until it reaches around 0.8 breaths/min, after which the predicted PERCLOS seems flat (see Figure \ref{fig:BRStd}). The predicted PERCLOS decreases when the value of br\_{avg60} increases from 1 beat/min to 2 beats/min. Then the trend tends to be reversed, i.e., the larger the value of  br\_{avg60}, the larger the value of the predicted PERCLOS (see Figure \ref{fig:HRStd}). Note the importance or the global impact of each individual feature is also noticeable in the range of predicted PERCLOS, where hr\_{avg60} has the maximum range, followed by heart\_{rate}\_{variability}, while the rest have similar ranges.  

\begin{figure}[H]
\centering
\includegraphics[width=\columnwidth]{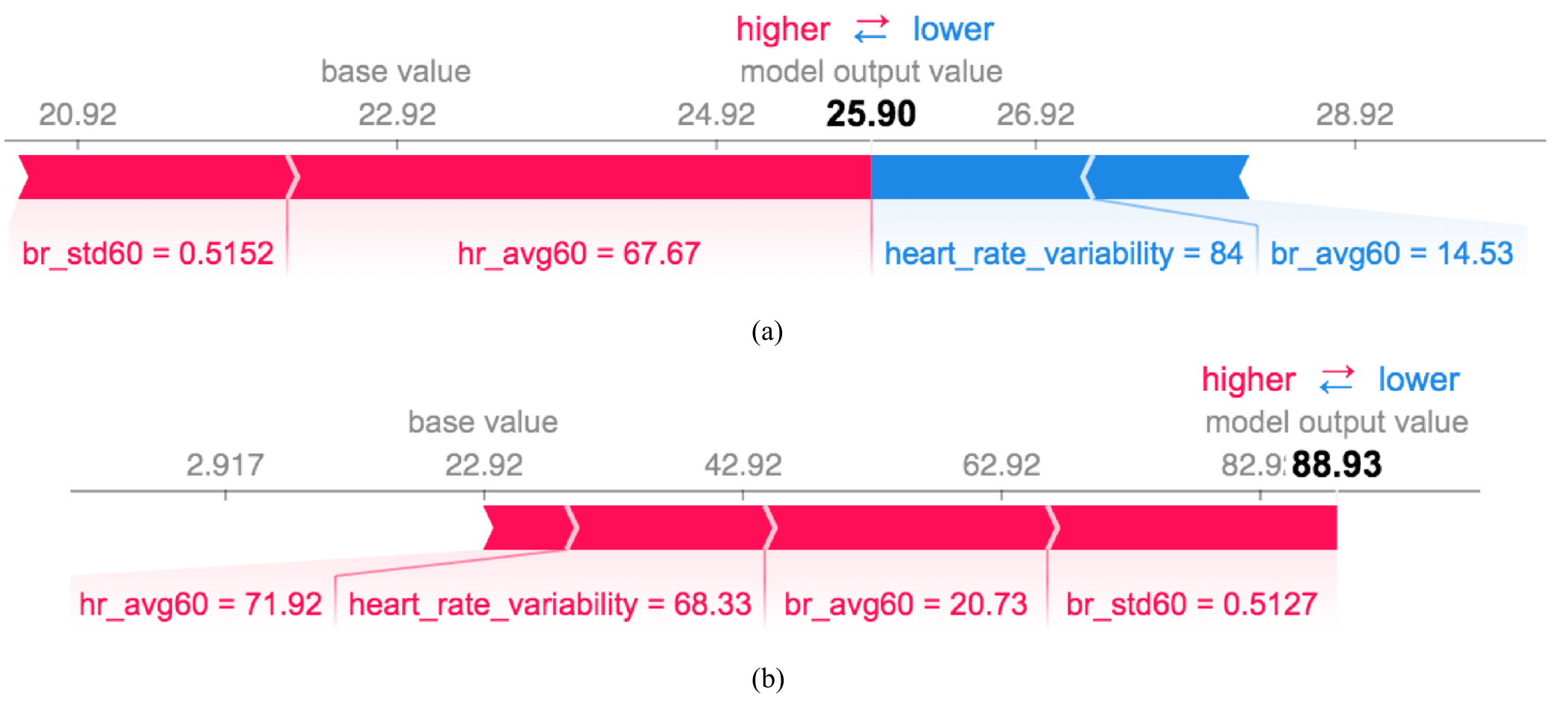}
\caption{SHAP individual explanations. (a) An example of low predicted driver fatigue; (b) An example of high predicted driver fatigue. Note the base value, 22.92, indicated the average predicted PERCLOS by all the training data in the model and those in blue push the predicted value lower and those in red push the predicted value higher.}
\label{fig:individual}
\end{figure}

\subsubsection{Individual Explanations}
SHAP is also able to produce individual explanations locally to show the contributions of each input feature. Figure \ref{fig:individual} shows two examples. The base value is the averaged output from the model, which was 22.92, the feature-value sets in blue push the output (predicted PERCLOS) lower, and the feature-value sets in red push the output higher. For example, the feature-value sets in Figure \ref{fig:individual}a push the output value to be 25.90, which was very close to the ground-truth value at 26.16. hr\_{avg60} and br\_{std60} push the predicted PERCLOS higher while br\_{avg60}, heart\_{rate}\_{variability} push the predicted PERCLOS lower. The feature hr\_{avg60} contributes the most to such an output. All the feature-value sets in Figure \ref{fig:individual}b push the output higher to be 88.93, which was also close to the ground true value at 90.91. Among them, br\_{std60} contributes the most, followed by br\_{avg60}, heart\_{rate}\_{variability}, and hr\_{avg60}. Therefore, despite the fact that globally hr\_{avg60} is the most important feature, its contribution might not always be the largest locally.     

\section{Discussions}
\subsection{Fatigue Prediction}
The proposed model used both physiological and behavioral measures to predict driver fatigue indicated by the PERCLOS measure. The model used 11 predictor variables as input and the model with best performance only made use of heart data and breathing data and was able to predict driver fatigue with high accuracy in real time. With Python 3.8 on a MacBook Pro with 2.3GHz Quad-Core Intel Core i7 and macOS Catalina, the average prediction time for one sample was only $3.6\times10^{-6} $ seconds. Compared with the fatigue-related studies in automated driving, our work used prediction models rather than simply described the fatigue progression in automated driving. Moreover, we identified the most important features in predicting driver fatigue in automated driving using SHAP, including hr\_{avg60}, heart\_{rate}\_{variability}, br\_{avg60}, br\_{std60}, and hr\_{std60}. Therefore, the included physiological measures were found to be more important than the included behavioral measures. The prediction model provided one good way to monitor, quantify, detect, and predict driver fatigue in real time. Despite the fact that fatigue has many components in terms of its bodily representation, multiple physiological measures were able to predict driver fatigue with RMSE = 3.847, MAE = 1.768, and adjusted $R^2 = 0.996$. During automated driving, wearable physiological sensors can be potentially used to detect and predict drivers’ fatigued state in real time in a minimally invasive manner. Such insights give us informed design guidelines in customizing driver fatigue models by tuning only the most critical physiological measures.

\subsection{Model Explanation}
Unlike previous driver fatigue prediction studies, the most important findings in this study are the relationships between the five most important measures identified by SHAP and driver fatigue indicated by PERCLOS. For example,  hr\_{avg60} and br\_{avg60} tended to be negatively correlated with predicted driver fatigue except at some specific, narrow intervals (see Figures \ref{fig:HR} and \ref{fig:BR}). This could be caused by the individual differences or noises involved in the dataset. However, such domain knowledge captured in the driver fatigue prediction model can be used to help design systems that fight driver fatigue in automated driving.

First, consistent with previous studies (e.g., \citep{unal2013driving}), a low heart rate can be indicative of a low level of arousal with low vigilance. In order to fight monotony in automated driving, music with high tempos, for example, can be used to increase drivers' heart rate to help drivers stay at an optimal level of arousal to improve driving performance \citep{dalton2007effects}. Compared to a control group with no music, participants with self-selected music increased 3 beats/min on average, which could decrease PERCLOS by as much as 40 (see  Figure \ref{fig:HR}). Second, consistent with previous studies (e.g., \citep{sun2011vehicle}), a decreasing breathing rate was also a sign of the onset of fatigue. To fight monotonous driver fatigue in automated driving, a breath booster system based on haptic guidance was proposed to increase breathing rate and heart rate in order to increase driver alertness and focus \citep{balters2018breath}. However, what is less known is that a smaller br\_{std60} was also associated with driver fatigue. A variable breath pattern could also be used to fight driver fatigue. Third, heart\_{rate}\_{variability} was computed as the standard deviation of inter-beat intervals while hr\_{std60} was calculated as the standard deviation of heart rate (see Table \ref{tab:predictor}). In addition, both had a V-shape relationship with the predicted PERCLOS (see Figures \ref{fig:HRV} and \ref{fig:HRStd}). In this sense, they described the same heart rate activity and its association with driver fatigue. Increases in heart\_{rate}\_{variability} could be associated with decreases in mental workload, which often occurred in sleepy drivers with monotonous driving \citep{horne1995sleep}. This was consistent with our finding when heart\_{rate}\_{variability} was between 50 and 140 ms or when hr\_{std60} was between 2 beats/min and 8 beats/min (see Figures \ref{fig:HRV} and \ref{fig:HRStd}). However, in other intervals, increases in heart\_{rate}\_{variability} led to decreases in driver fatigue, especially when it was smaller than 50 ms or hr\_{std60} was smaller than 2 beats/min. This was not reported previously and could be potentially explained by the different measures used for heart rate variability. For example, \citet{fujiwara2018heart} specifically included a feature named NN50, which was defined as the number of adjacent inter-beat intervals whose difference was more than 50 ms within a period of time. This is consistent with our results, where 50 ms was the turning point in the V-shape relationship between heart\_{rate}\_{variability} and the predicted PERCLOS. 

\subsection{Limitations and Future Work}
First, we used XGBoost to predict driver fatigue without considering the temporal relationships among the training data. A model, such as LSTM, can potentially improve the performance of the model further by examining the temporal relationships in the data in the future. However, the cost of better performance of LSTM is that it would be difficult to explain the captured knowledge by LSTM using SHAP. Moreover, it might still be not adequate to detect and predict driver fatigue in real time and more research should be devoted to predicting driver fatigue ahead of time in order for the driver to prepare possible hazards in the takeover process in automated driving \citep{zhou2020driver}. Second, it should be cautious to generalize our results to other situations because driver fatigue in this study mainly refers to passive fatigue due to monotonous automated driving, which can be different from fatigue caused by sleep-deprivation in traditional manual driving. In this sense, our model is more appropriate for fatigue monitoring in automated driving rather than in manual driving. 

\section{Conclusion}
In this study, we built a fatigue prediction model using XGBoost in automated driving. In order to understand the black-box XGBoost model, we utilized SHAP based on coalitional game theory. First, SHAP was used to identify the most important measures among the 11 predictor variables and using only the top five most important predictor variables, the XGBoost was able to predict driver fatigue indicated by PERCLOS accurately. Second, SHAP was able to identify the main effects of the important predictor variables in the XGBoost model globally. Third, SHAP also offered individual prediction explanations to understand the contributions of each predictor variable locally. These insights can potentially help automotive manufacturers design more acceptable and trustworthy fatigue detection and prediction models in automated vehicles.

\bibliography{main}

\begin{thebibliography}{}

\bibitem [\protect \citeauthoryear {%
Ayoub%
, Yang%
\BCBL {}\ \BBA {} Zhou%
}{%
Ayoub%
\ \protect \BOthers {.}}{%
{\protect \APACyear {2021}}%
}]{%
AYOUB2021102}
\APACinsertmetastar {%
AYOUB2021102}%
\begin{APACrefauthors}%
Ayoub, J.%
, Yang, X\BPBI J.%
\BCBL {}\ \BBA {} Zhou, F.%
\end{APACrefauthors}%
\unskip\
\newblock
\APACrefYearMonthDay{2021}{}{}.
\newblock
{\BBOQ}\APACrefatitle {Modeling dispositional and initial learned trust in
  automated vehicles with predictability and explainability} {Modeling
  dispositional and initial learned trust in automated vehicles with
  predictability and explainability}.{\BBCQ}
\newblock
\APACjournalVolNumPages{Transportation Research Part F: Traffic Psychology and
  Behaviour}{77}{}{102 - 116}.
\PrintBackRefs{\CurrentBib}

\bibitem [\protect \citeauthoryear {%
Ayoub%
, Zhou%
, Bao%
\BCBL {}\ \BBA {} Yang%
}{%
Ayoub%
\ \protect \BOthers {.}}{%
{\protect \APACyear {2019}}%
}]{%
ayoub2019manual}
\APACinsertmetastar {%
ayoub2019manual}%
\begin{APACrefauthors}%
Ayoub, J.%
, Zhou, F.%
, Bao, S.%
\BCBL {}\ \BBA {} Yang, X\BPBI J.%
\end{APACrefauthors}%
\unskip\
\newblock
\APACrefYearMonthDay{2019}{}{}.
\newblock
{\BBOQ}\APACrefatitle {From manual driving to automated driving: A review of 10
  years of autoui} {From manual driving to automated driving: A review of 10
  years of autoui}.{\BBCQ}
\newblock
\BIn{} \APACrefbtitle {Proceedings of the 11th International Conference on
  Automotive User Interfaces and Interactive Vehicular Applications}
  {Proceedings of the 11th international conference on automotive user
  interfaces and interactive vehicular applications}\ (\BPGS\ 70--90).
\PrintBackRefs{\CurrentBib}

\bibitem [\protect \citeauthoryear {%
Balters%
, Murnane%
, Landay%
\BCBL {}\ \BBA {} Paredes%
}{%
Balters%
\ \protect \BOthers {.}}{%
{\protect \APACyear {2018}}%
}]{%
balters2018breath}
\APACinsertmetastar {%
balters2018breath}%
\begin{APACrefauthors}%
Balters, S.%
, Murnane, E\BPBI L.%
, Landay, J\BPBI A.%
\BCBL {}\ \BBA {} Paredes, P\BPBI E.%
\end{APACrefauthors}%
\unskip\
\newblock
\APACrefYearMonthDay{2018}{}{}.
\newblock
{\BBOQ}\APACrefatitle {Breath Booster! Exploring In-car, Fast-paced Breathing
  Interventions to Enhance Driver Arousal State} {Breath booster! exploring
  in-car, fast-paced breathing interventions to enhance driver arousal
  state}.{\BBCQ}
\newblock
\BIn{} \APACrefbtitle {Proceedings of the 12th EAI International Conference on
  Pervasive Computing Technologies for Healthcare} {Proceedings of the 12th eai
  international conference on pervasive computing technologies for healthcare}\
  (\BPGS\ 128--137).
\PrintBackRefs{\CurrentBib}

\bibitem [\protect \citeauthoryear {%
Caruana%
\ \protect \BOthers {.}}{%
Caruana%
\ \protect \BOthers {.}}{%
{\protect \APACyear {2015}}%
}]{%
caruana2015intelligible}
\APACinsertmetastar {%
caruana2015intelligible}%
\begin{APACrefauthors}%
Caruana, R.%
, Lou, Y.%
, Gehrke, J.%
, Koch, P.%
, Sturm, M.%
\BCBL {}\ \BBA {} Elhadad, N.%
\end{APACrefauthors}%
\unskip\
\newblock
\APACrefYearMonthDay{2015}{}{}.
\newblock
{\BBOQ}\APACrefatitle {Intelligible models for healthcare: Predicting pneumonia
  risk and hospital 30-day readmission} {Intelligible models for healthcare:
  Predicting pneumonia risk and hospital 30-day readmission}.{\BBCQ}
\newblock
\BIn{} \APACrefbtitle {Proceedings of the 21th ACM SIGKDD international
  conference on knowledge discovery and data mining} {Proceedings of the 21th
  acm sigkdd international conference on knowledge discovery and data mining}\
  (\BPGS\ 1721--1730).
\PrintBackRefs{\CurrentBib}

\bibitem [\protect \citeauthoryear {%
Chang%
\ \BBA {} Chen%
}{%
Chang%
\ \BBA {} Chen%
}{%
{\protect \APACyear {2014}}%
}]{%
chang2014driver}
\APACinsertmetastar {%
chang2014driver}%
\begin{APACrefauthors}%
Chang, T\BHBI H.%
\BCBT {}\ \BBA {} Chen, Y\BHBI R.%
\end{APACrefauthors}%
\unskip\
\newblock
\APACrefYearMonthDay{2014}{}{}.
\newblock
{\BBOQ}\APACrefatitle {Driver fatigue surveillance via eye detection} {Driver
  fatigue surveillance via eye detection}.{\BBCQ}
\newblock
\BIn{} \APACrefbtitle {17th International IEEE Conference on Intelligent
  Transportation Systems (ITSC)} {17th international ieee conference on
  intelligent transportation systems (itsc)}\ (\BPGS\ 366--371).
\PrintBackRefs{\CurrentBib}

\bibitem [\protect \citeauthoryear {%
Chen%
\ \BBA {} Guestrin%
}{%
Chen%
\ \BBA {} Guestrin%
}{%
{\protect \APACyear {2016}}%
}]{%
chen2016xgboost}
\APACinsertmetastar {%
chen2016xgboost}%
\begin{APACrefauthors}%
Chen, T.%
\BCBT {}\ \BBA {} Guestrin, C.%
\end{APACrefauthors}%
\unskip\
\newblock
\APACrefYearMonthDay{2016}{}{}.
\newblock
{\BBOQ}\APACrefatitle {Xgboost: A scalable tree boosting system} {Xgboost: A
  scalable tree boosting system}.{\BBCQ}
\newblock
\BIn{} \APACrefbtitle {Proceedings of the 22nd acm sigkdd international
  conference on knowledge discovery and data mining} {Proceedings of the 22nd
  acm sigkdd international conference on knowledge discovery and data mining}\
  (\BPGS\ 785--794).
\PrintBackRefs{\CurrentBib}

\bibitem [\protect \citeauthoryear {%
Clark%
, McLaughlin%
, Williams%
\BCBL {}\ \BBA {} Feng%
}{%
Clark%
\ \protect \BOthers {.}}{%
{\protect \APACyear {2017}}%
}]{%
clark2017performance}
\APACinsertmetastar {%
clark2017performance}%
\begin{APACrefauthors}%
Clark, H.%
, McLaughlin, A\BPBI C.%
, Williams, B.%
\BCBL {}\ \BBA {} Feng, J.%
\end{APACrefauthors}%
\unskip\
\newblock
\APACrefYearMonthDay{2017}{}{}.
\newblock
{\BBOQ}\APACrefatitle {Performance in takeover and characteristics of
  non-driving related tasks during highly automated driving in younger and
  older drivers} {Performance in takeover and characteristics of non-driving
  related tasks during highly automated driving in younger and older
  drivers}.{\BBCQ}
\newblock
\BIn{} \APACrefbtitle {Proceedings of the Human Factors and Ergonomics Society
  Annual Meeting} {Proceedings of the human factors and ergonomics society
  annual meeting}\ (\BVOL~61, \BPGS\ 37--41).
\PrintBackRefs{\CurrentBib}

\bibitem [\protect \citeauthoryear {%
Collet%
\ \BBA {} Musicant%
}{%
Collet%
\ \BBA {} Musicant%
}{%
{\protect \APACyear {2019}}%
}]{%
collet2019associating}
\APACinsertmetastar {%
collet2019associating}%
\begin{APACrefauthors}%
Collet, C.%
\BCBT {}\ \BBA {} Musicant, O.%
\end{APACrefauthors}%
\unskip\
\newblock
\APACrefYearMonthDay{2019}{}{}.
\newblock
{\BBOQ}\APACrefatitle {Associating vehicles automation with drivers functional
  state assessment systems: A challenge for road safety in the future}
  {Associating vehicles automation with drivers functional state assessment
  systems: A challenge for road safety in the future}.{\BBCQ}
\newblock
\APACjournalVolNumPages{Frontiers in human neuroscience}{13}{}{131}.
\PrintBackRefs{\CurrentBib}

\bibitem [\protect \citeauthoryear {%
Dalton%
, Behm%
\BCBL {}\ \BBA {} Kibele%
}{%
Dalton%
\ \protect \BOthers {.}}{%
{\protect \APACyear {2007}}%
}]{%
dalton2007effects}
\APACinsertmetastar {%
dalton2007effects}%
\begin{APACrefauthors}%
Dalton, B\BPBI H.%
, Behm, D\BPBI G.%
\BCBL {}\ \BBA {} Kibele, A.%
\end{APACrefauthors}%
\unskip\
\newblock
\APACrefYearMonthDay{2007}{}{}.
\newblock
{\BBOQ}\APACrefatitle {Effects of sound types and volumes on simulated driving,
  vigilance tasks and heart rate} {Effects of sound types and volumes on
  simulated driving, vigilance tasks and heart rate}.{\BBCQ}
\newblock
\APACjournalVolNumPages{Occupational Ergonomics}{7}{3}{153--168}.
\PrintBackRefs{\CurrentBib}

\bibitem [\protect \citeauthoryear {%
Dong%
, Hu%
, Uchimura%
\BCBL {}\ \BBA {} Murayama%
}{%
Dong%
\ \protect \BOthers {.}}{%
{\protect \APACyear {2010}}%
}]{%
dong2010driver}
\APACinsertmetastar {%
dong2010driver}%
\begin{APACrefauthors}%
Dong, Y.%
, Hu, Z.%
, Uchimura, K.%
\BCBL {}\ \BBA {} Murayama, N.%
\end{APACrefauthors}%
\unskip\
\newblock
\APACrefYearMonthDay{2010}{}{}.
\newblock
{\BBOQ}\APACrefatitle {Driver inattention monitoring system for intelligent
  vehicles: A review} {Driver inattention monitoring system for intelligent
  vehicles: A review}.{\BBCQ}
\newblock
\APACjournalVolNumPages{IEEE transactions on intelligent transportation
  systems}{12}{2}{596--614}.
\PrintBackRefs{\CurrentBib}

\bibitem [\protect \citeauthoryear {%
Doshi-Velez%
\ \BBA {} Kim%
}{%
Doshi-Velez%
\ \BBA {} Kim%
}{%
{\protect \APACyear {2017}}%
}]{%
doshi2017towards}
\APACinsertmetastar {%
doshi2017towards}%
\begin{APACrefauthors}%
Doshi-Velez, F.%
\BCBT {}\ \BBA {} Kim, B.%
\end{APACrefauthors}%
\unskip\
\newblock
\APACrefYearMonthDay{2017}{}{}.
\newblock
{\BBOQ}\APACrefatitle {Towards a rigorous science of interpretable machine
  learning} {Towards a rigorous science of interpretable machine
  learning}.{\BBCQ}
\newblock
\APACjournalVolNumPages{arXiv preprint arXiv:1702.08608}{}{}{}.
\PrintBackRefs{\CurrentBib}

\bibitem [\protect \citeauthoryear {%
Du%
, Yang%
\BCBL {}\ \BBA {} Zhou%
}{%
Du%
, Yang%
\BCBL {}\ \BBA {} Zhou%
}{%
{\protect \APACyear {2020}}%
}]{%
du2020psychophysiological}
\APACinsertmetastar {%
du2020psychophysiological}%
\begin{APACrefauthors}%
Du, N.%
, Yang, X\BPBI J.%
\BCBL {}\ \BBA {} Zhou, F.%
\end{APACrefauthors}%
\unskip\
\newblock
\APACrefYearMonthDay{2020}{}{}.
\newblock
{\BBOQ}\APACrefatitle {Psychophysiological responses to takeover requests in
  conditionally automated driving} {Psychophysiological responses to takeover
  requests in conditionally automated driving}.{\BBCQ}
\newblock
\APACjournalVolNumPages{Accident Analysis \& Prevention}{148}{}{105804}.
\PrintBackRefs{\CurrentBib}

\bibitem [\protect \citeauthoryear {%
Du%
, Zhou%
\BCBL {}\ \protect \BOthers {.}}{%
Du%
, Zhou%
\BCBL {}\ \protect \BOthers {.}}{%
{\protect \APACyear {2020}}%
{\protect \APACexlab {{\protect \BCnt {3}}}}}]{%
du2020predicting1}
\APACinsertmetastar {%
du2020predicting1}%
\begin{APACrefauthors}%
Du, N.%
, Zhou, F.%
, Pulver, E.%
, Tilbury, D.%
, Robert, L\BPBI P.%
, Pradhan, A\BPBI K.%
\BCBL {}\ \BBA {} Yang, X\BPBI J.%
\end{APACrefauthors}%
\unskip\
\newblock
\APACrefYearMonthDay{2020{\protect \BCnt {3}}}{}{}.
\newblock
{\BBOQ}\APACrefatitle {Predicting Takeover Performance in Conditionally
  Automated Driving} {Predicting takeover performance in conditionally
  automated driving}.{\BBCQ}
\newblock
\BIn{} \APACrefbtitle {Extended Abstracts of the 2020 CHI Conference on Human
  Factors in Computing Systems} {Extended abstracts of the 2020 chi conference
  on human factors in computing systems}\ (\BPGS\ 1--8).
\PrintBackRefs{\CurrentBib}

\bibitem [\protect \citeauthoryear {%
Du%
, Zhou%
\BCBL {}\ \protect \BOthers {.}}{%
Du%
, Zhou%
\BCBL {}\ \protect \BOthers {.}}{%
{\protect \APACyear {2020}}%
{\protect \APACexlab {{\protect \BCnt {1}}}}}]{%
du2020examining}
\APACinsertmetastar {%
du2020examining}%
\begin{APACrefauthors}%
Du, N.%
, Zhou, F.%
, Pulver, E\BPBI M.%
, Tilbury, D\BPBI M.%
, Robert, L\BPBI P.%
, Pradhan, A\BPBI K.%
\BCBL {}\ \BBA {} Yang, X\BPBI J.%
\end{APACrefauthors}%
\unskip\
\newblock
\APACrefYearMonthDay{2020{\protect \BCnt {1}}}{}{}.
\newblock
{\BBOQ}\APACrefatitle {Examining the effects of emotional valence and arousal
  on takeover performance in conditionally automated driving} {Examining the
  effects of emotional valence and arousal on takeover performance in
  conditionally automated driving}.{\BBCQ}
\newblock
\APACjournalVolNumPages{Transportation research part C: emerging
  technologies}{112}{}{78--87}.
\PrintBackRefs{\CurrentBib}

\bibitem [\protect \citeauthoryear {%
Du%
, Zhou%
\BCBL {}\ \protect \BOthers {.}}{%
Du%
, Zhou%
\BCBL {}\ \protect \BOthers {.}}{%
{\protect \APACyear {2020}}%
{\protect \APACexlab {{\protect \BCnt {2}}}}}]{%
du2020predicting}
\APACinsertmetastar {%
du2020predicting}%
\begin{APACrefauthors}%
Du, N.%
, Zhou, F.%
, Pulver, E\BPBI M.%
, Tilbury, D\BPBI M.%
, Robert, L\BPBI P.%
, Pradhan, A\BPBI K.%
\BCBL {}\ \BBA {} Yang, X\BPBI J.%
\end{APACrefauthors}%
\unskip\
\newblock
\APACrefYearMonthDay{2020{\protect \BCnt {2}}}{}{}.
\newblock
{\BBOQ}\APACrefatitle {Predicting driver takeover performance in conditionally
  automated driving} {Predicting driver takeover performance in conditionally
  automated driving}.{\BBCQ}
\newblock
\APACjournalVolNumPages{Accident Analysis \& Prevention}{148}{}{105748}.
\PrintBackRefs{\CurrentBib}

\bibitem [\protect \citeauthoryear {%
Dwivedi%
, Biswaranjan%
\BCBL {}\ \BBA {} Sethi%
}{%
Dwivedi%
\ \protect \BOthers {.}}{%
{\protect \APACyear {2014}}%
}]{%
dwivedi2014drowsy}
\APACinsertmetastar {%
dwivedi2014drowsy}%
\begin{APACrefauthors}%
Dwivedi, K.%
, Biswaranjan, K.%
\BCBL {}\ \BBA {} Sethi, A.%
\end{APACrefauthors}%
\unskip\
\newblock
\APACrefYearMonthDay{2014}{}{}.
\newblock
{\BBOQ}\APACrefatitle {Drowsy driver detection using representation learning}
  {Drowsy driver detection using representation learning}.{\BBCQ}
\newblock
\BIn{} \APACrefbtitle {2014 IEEE international advance computing conference
  (IACC)} {2014 ieee international advance computing conference (iacc)}\
  (\BPGS\ 995--999).
\PrintBackRefs{\CurrentBib}

\bibitem [\protect \citeauthoryear {%
Feldh{\"u}tter%
, Gold%
, Schneider%
\BCBL {}\ \BBA {} Bengler%
}{%
Feldh{\"u}tter%
\ \protect \BOthers {.}}{%
{\protect \APACyear {2017}}%
}]{%
feldhutter2017duration}
\APACinsertmetastar {%
feldhutter2017duration}%
\begin{APACrefauthors}%
Feldh{\"u}tter, A.%
, Gold, C.%
, Schneider, S.%
\BCBL {}\ \BBA {} Bengler, K.%
\end{APACrefauthors}%
\unskip\
\newblock
\APACrefYearMonthDay{2017}{}{}.
\newblock
{\BBOQ}\APACrefatitle {How the duration of automated driving influences
  take-over performance and gaze behavior} {How the duration of automated
  driving influences take-over performance and gaze behavior}.{\BBCQ}
\newblock
\BIn{} \APACrefbtitle {Advances in ergonomic design of systems, products and
  processes} {Advances in ergonomic design of systems, products and processes}\
  (\BPGS\ 309--318).
\newblock
\APACaddressPublisher{}{Springer}.
\PrintBackRefs{\CurrentBib}

\bibitem [\protect \citeauthoryear {%
Feng%
, Zhang%
\BCBL {}\ \BBA {} Cheng%
}{%
Feng%
\ \protect \BOthers {.}}{%
{\protect \APACyear {2009}}%
}]{%
feng2009board}
\APACinsertmetastar {%
feng2009board}%
\begin{APACrefauthors}%
Feng, R.%
, Zhang, G.%
\BCBL {}\ \BBA {} Cheng, B.%
\end{APACrefauthors}%
\unskip\
\newblock
\APACrefYearMonthDay{2009}{}{}.
\newblock
{\BBOQ}\APACrefatitle {An on-board system for detecting driver drowsiness based
  on multi-sensor data fusion using Dempster-Shafer theory} {An on-board system
  for detecting driver drowsiness based on multi-sensor data fusion using
  dempster-shafer theory}.{\BBCQ}
\newblock
\BIn{} \APACrefbtitle {2009 International Conference on Networking, Sensing and
  Control} {2009 international conference on networking, sensing and control}\
  (\BPGS\ 897--902).
\PrintBackRefs{\CurrentBib}

\bibitem [\protect \citeauthoryear {%
Fujiwara%
\ \protect \BOthers {.}}{%
Fujiwara%
\ \protect \BOthers {.}}{%
{\protect \APACyear {2018}}%
}]{%
fujiwara2018heart}
\APACinsertmetastar {%
fujiwara2018heart}%
\begin{APACrefauthors}%
Fujiwara, K.%
, Abe, E.%
, Kamata, K.%
, Nakayama, C.%
, Suzuki, Y.%
, Yamakawa, T.%
\BDBL {}others%
\end{APACrefauthors}%
\unskip\
\newblock
\APACrefYearMonthDay{2018}{}{}.
\newblock
{\BBOQ}\APACrefatitle {Heart rate variability-based driver drowsiness detection
  and its validation with EEG} {Heart rate variability-based driver drowsiness
  detection and its validation with eeg}.{\BBCQ}
\newblock
\APACjournalVolNumPages{IEEE Transactions on Biomedical
  Engineering}{66}{6}{1769--1778}.
\PrintBackRefs{\CurrentBib}

\bibitem [\protect \citeauthoryear {%
Gon{\c{c}}alves%
, Happee%
\BCBL {}\ \BBA {} Bengler%
}{%
Gon{\c{c}}alves%
\ \protect \BOthers {.}}{%
{\protect \APACyear {2016}}%
}]{%
gonccalves2016drowsiness}
\APACinsertmetastar {%
gonccalves2016drowsiness}%
\begin{APACrefauthors}%
Gon{\c{c}}alves, J.%
, Happee, R.%
\BCBL {}\ \BBA {} Bengler, K.%
\end{APACrefauthors}%
\unskip\
\newblock
\APACrefYearMonthDay{2016}{}{}.
\newblock
{\BBOQ}\APACrefatitle {Drowsiness in conditional automation: proneness,
  diagnosis and driving performance effects} {Drowsiness in conditional
  automation: proneness, diagnosis and driving performance effects}.{\BBCQ}
\newblock
\BIn{} \APACrefbtitle {2016 IEEE 19th international conference on intelligent
  transportation systems (ITSC)} {2016 ieee 19th international conference on
  intelligent transportation systems (itsc)}\ (\BPGS\ 873--878).
\PrintBackRefs{\CurrentBib}

\bibitem [\protect \citeauthoryear {%
Hadi%
, Li%
, Wang%
, Yuan%
\BCBL {}\ \BBA {} Cheng%
}{%
Hadi%
\ \protect \BOthers {.}}{%
{\protect \APACyear {2020}}%
}]{%
hadi2020influence}
\APACinsertmetastar {%
hadi2020influence}%
\begin{APACrefauthors}%
Hadi, A\BPBI M.%
, Li, Q.%
, Wang, W.%
, Yuan, Q.%
\BCBL {}\ \BBA {} Cheng, B.%
\end{APACrefauthors}%
\unskip\
\newblock
\APACrefYearMonthDay{2020}{}{}.
\newblock
{\BBOQ}\APACrefatitle {Influence of Passive Fatigue and Take-Over Request Lead
  Time on Drivers’ Take-Over Performance} {Influence of passive fatigue and
  take-over request lead time on drivers’ take-over performance}.{\BBCQ}
\newblock
\BIn{} \APACrefbtitle {International Conference on Applied Human Factors and
  Ergonomics} {International conference on applied human factors and
  ergonomics}\ (\BPGS\ 253--259).
\PrintBackRefs{\CurrentBib}

\bibitem [\protect \citeauthoryear {%
Horne%
\ \BBA {} Reyner%
}{%
Horne%
\ \BBA {} Reyner%
}{%
{\protect \APACyear {1995}}%
}]{%
horne1995sleep}
\APACinsertmetastar {%
horne1995sleep}%
\begin{APACrefauthors}%
Horne, J\BPBI A.%
\BCBT {}\ \BBA {} Reyner, L\BPBI A.%
\end{APACrefauthors}%
\unskip\
\newblock
\APACrefYearMonthDay{1995}{}{}.
\newblock
{\BBOQ}\APACrefatitle {Sleep related vehicle accidents} {Sleep related vehicle
  accidents}.{\BBCQ}
\newblock
\APACjournalVolNumPages{Bmj}{310}{6979}{565--567}.
\PrintBackRefs{\CurrentBib}

\bibitem [\protect \citeauthoryear {%
Ji%
, Zhu%
\BCBL {}\ \BBA {} Lan%
}{%
Ji%
\ \protect \BOthers {.}}{%
{\protect \APACyear {2004}}%
}]{%
ji2004real}
\APACinsertmetastar {%
ji2004real}%
\begin{APACrefauthors}%
Ji, Q.%
, Zhu, Z.%
\BCBL {}\ \BBA {} Lan, P.%
\end{APACrefauthors}%
\unskip\
\newblock
\APACrefYearMonthDay{2004}{}{}.
\newblock
{\BBOQ}\APACrefatitle {Real-time nonintrusive monitoring and prediction of
  driver fatigue} {Real-time nonintrusive monitoring and prediction of driver
  fatigue}.{\BBCQ}
\newblock
\APACjournalVolNumPages{IEEE transactions on vehicular
  technology}{53}{4}{1052--1068}.
\PrintBackRefs{\CurrentBib}

\bibitem [\protect \citeauthoryear {%
Jung%
, Shin%
\BCBL {}\ \BBA {} Chung%
}{%
Jung%
\ \protect \BOthers {.}}{%
{\protect \APACyear {2014}}%
}]{%
jung2014driver}
\APACinsertmetastar {%
jung2014driver}%
\begin{APACrefauthors}%
Jung, S\BHBI J.%
, Shin, H\BHBI S.%
\BCBL {}\ \BBA {} Chung, W\BHBI Y.%
\end{APACrefauthors}%
\unskip\
\newblock
\APACrefYearMonthDay{2014}{}{}.
\newblock
{\BBOQ}\APACrefatitle {Driver fatigue and drowsiness monitoring system with
  embedded electrocardiogram sensor on steering wheel} {Driver fatigue and
  drowsiness monitoring system with embedded electrocardiogram sensor on
  steering wheel}.{\BBCQ}
\newblock
\APACjournalVolNumPages{IET Intelligent Transport Systems}{8}{1}{43--50}.
\PrintBackRefs{\CurrentBib}

\bibitem [\protect \citeauthoryear {%
Khan%
\ \BBA {} Mansoor%
}{%
Khan%
\ \BBA {} Mansoor%
}{%
{\protect \APACyear {2008}}%
}]{%
khan2008real}
\APACinsertmetastar {%
khan2008real}%
\begin{APACrefauthors}%
Khan, M\BPBI I.%
\BCBT {}\ \BBA {} Mansoor, A\BPBI B.%
\end{APACrefauthors}%
\unskip\
\newblock
\APACrefYearMonthDay{2008}{}{}.
\newblock
{\BBOQ}\APACrefatitle {Real time eyes tracking and classification for driver
  fatigue detection} {Real time eyes tracking and classification for driver
  fatigue detection}.{\BBCQ}
\newblock
\BIn{} \APACrefbtitle {International Conference Image Analysis and Recognition}
  {International conference image analysis and recognition}\ (\BPGS\ 729--738).
\PrintBackRefs{\CurrentBib}

\bibitem [\protect \citeauthoryear {%
Kher%
}{%
Kher%
}{%
{\protect \APACyear {2019}}%
}]{%
kher2019signal}
\APACinsertmetastar {%
kher2019signal}%
\begin{APACrefauthors}%
Kher, R.%
\end{APACrefauthors}%
\unskip\
\newblock
\APACrefYearMonthDay{2019}{}{}.
\newblock
{\BBOQ}\APACrefatitle {Signal processing techniques for removing noise from ECG
  signals} {Signal processing techniques for removing noise from ecg
  signals}.{\BBCQ}
\newblock
\APACjournalVolNumPages{J. Biomed. Eng. Res}{3}{}{1--9}.
\PrintBackRefs{\CurrentBib}

\bibitem [\protect \citeauthoryear {%
Koesdwiady%
, Soua%
, Karray%
\BCBL {}\ \BBA {} Kamel%
}{%
Koesdwiady%
\ \protect \BOthers {.}}{%
{\protect \APACyear {2016}}%
}]{%
koesdwiady2016recent}
\APACinsertmetastar {%
koesdwiady2016recent}%
\begin{APACrefauthors}%
Koesdwiady, A.%
, Soua, R.%
, Karray, F.%
\BCBL {}\ \BBA {} Kamel, M\BPBI S.%
\end{APACrefauthors}%
\unskip\
\newblock
\APACrefYearMonthDay{2016}{}{}.
\newblock
{\BBOQ}\APACrefatitle {Recent trends in driver safety monitoring systems: State
  of the art and challenges} {Recent trends in driver safety monitoring
  systems: State of the art and challenges}.{\BBCQ}
\newblock
\APACjournalVolNumPages{IEEE transactions on vehicular
  technology}{66}{6}{4550--4563}.
\PrintBackRefs{\CurrentBib}

\bibitem [\protect \citeauthoryear {%
K{\"o}rber%
, Cingel%
, Zimmermann%
\BCBL {}\ \BBA {} Bengler%
}{%
K{\"o}rber%
\ \protect \BOthers {.}}{%
{\protect \APACyear {2015}}%
}]{%
korber2015vigilance}
\APACinsertmetastar {%
korber2015vigilance}%
\begin{APACrefauthors}%
K{\"o}rber, M.%
, Cingel, A.%
, Zimmermann, M.%
\BCBL {}\ \BBA {} Bengler, K.%
\end{APACrefauthors}%
\unskip\
\newblock
\APACrefYearMonthDay{2015}{}{}.
\newblock
{\BBOQ}\APACrefatitle {Vigilance decrement and passive fatigue caused by
  monotony in automated driving} {Vigilance decrement and passive fatigue
  caused by monotony in automated driving}.{\BBCQ}
\newblock
\APACjournalVolNumPages{Procedia Manufacturing}{3}{}{2403--2409}.
\PrintBackRefs{\CurrentBib}

\bibitem [\protect \citeauthoryear {%
Krajewski%
, Sommer%
, Trutschel%
, Edwards%
\BCBL {}\ \BBA {} Golz%
}{%
Krajewski%
\ \protect \BOthers {.}}{%
{\protect \APACyear {2009}}%
}]{%
krajewski2009steering}
\APACinsertmetastar {%
krajewski2009steering}%
\begin{APACrefauthors}%
Krajewski, J.%
, Sommer, D.%
, Trutschel, U.%
, Edwards, D.%
\BCBL {}\ \BBA {} Golz, M.%
\end{APACrefauthors}%
\unskip\
\newblock
\APACrefYearMonthDay{2009}{}{}.
\newblock
{\BBOQ}\APACrefatitle {Steering wheel behavior based estimation of fatigue}
  {Steering wheel behavior based estimation of fatigue}.{\BBCQ}
\newblock
\BIn{} \APACrefbtitle {Proceedings of the... international driving symposium on
  human factors in driver assessment, training and vehicle design} {Proceedings
  of the... international driving symposium on human factors in driver
  assessment, training and vehicle design}\ (\BVOL~5, \BPGS\ 118--124).
\PrintBackRefs{\CurrentBib}

\bibitem [\protect \citeauthoryear {%
Kumar%
, Kalia%
\BCBL {}\ \BBA {} Sharma%
}{%
Kumar%
\ \protect \BOthers {.}}{%
{\protect \APACyear {2017}}%
}]{%
kumar2017predictive}
\APACinsertmetastar {%
kumar2017predictive}%
\begin{APACrefauthors}%
Kumar, S.%
, Kalia, A.%
\BCBL {}\ \BBA {} Sharma, A.%
\end{APACrefauthors}%
\unskip\
\newblock
\APACrefYearMonthDay{2017}{}{}.
\newblock
{\BBOQ}\APACrefatitle {Predictive analysis of alertness related features for
  driver drowsiness detection} {Predictive analysis of alertness related
  features for driver drowsiness detection}.{\BBCQ}
\newblock
\BIn{} \APACrefbtitle {International Conference on Intelligent Systems Design
  and Applications} {International conference on intelligent systems design and
  applications}\ (\BPGS\ 368--377).
\PrintBackRefs{\CurrentBib}

\bibitem [\protect \citeauthoryear {%
Lee%
\ \BBA {} Chung%
}{%
Lee%
\ \BBA {} Chung%
}{%
{\protect \APACyear {2012}}%
}]{%
lee2012driver}
\APACinsertmetastar {%
lee2012driver}%
\begin{APACrefauthors}%
Lee, B\BHBI G.%
\BCBT {}\ \BBA {} Chung, W\BHBI Y.%
\end{APACrefauthors}%
\unskip\
\newblock
\APACrefYearMonthDay{2012}{}{}.
\newblock
{\BBOQ}\APACrefatitle {Driver alertness monitoring using fusion of facial
  features and bio-signals} {Driver alertness monitoring using fusion of facial
  features and bio-signals}.{\BBCQ}
\newblock
\APACjournalVolNumPages{IEEE Sensors Journal}{12}{7}{2416--2422}.
\PrintBackRefs{\CurrentBib}

\bibitem [\protect \citeauthoryear {%
Li%
, Chen%
, Peng%
\BCBL {}\ \BBA {} Wu%
}{%
Li%
\ \protect \BOthers {.}}{%
{\protect \APACyear {2017}}%
}]{%
li2017automatic}
\APACinsertmetastar {%
li2017automatic}%
\begin{APACrefauthors}%
Li, Z.%
, Chen, L.%
, Peng, J.%
\BCBL {}\ \BBA {} Wu, Y.%
\end{APACrefauthors}%
\unskip\
\newblock
\APACrefYearMonthDay{2017}{}{}.
\newblock
{\BBOQ}\APACrefatitle {Automatic detection of driver fatigue using driving
  operation information for transportation safety} {Automatic detection of
  driver fatigue using driving operation information for transportation
  safety}.{\BBCQ}
\newblock
\APACjournalVolNumPages{Sensors}{17}{6}{1212}.
\PrintBackRefs{\CurrentBib}

\bibitem [\protect \citeauthoryear {%
Lundberg%
\ \protect \BOthers {.}}{%
Lundberg%
\ \protect \BOthers {.}}{%
{\protect \APACyear {2020}}%
}]{%
lundberg2020local}
\APACinsertmetastar {%
lundberg2020local}%
\begin{APACrefauthors}%
Lundberg, S\BPBI M.%
, Erion, G.%
, Chen, H.%
, DeGrave, A.%
, Prutkin, J\BPBI M.%
, Nair, B.%
\BDBL {}Lee, S\BHBI I.%
\end{APACrefauthors}%
\unskip\
\newblock
\APACrefYearMonthDay{2020}{}{}.
\newblock
{\BBOQ}\APACrefatitle {From local explanations to global understanding with
  explainable AI for trees} {From local explanations to global understanding
  with explainable ai for trees}.{\BBCQ}
\newblock
\APACjournalVolNumPages{Nature machine intelligence}{2}{1}{2522--5839}.
\PrintBackRefs{\CurrentBib}

\bibitem [\protect \citeauthoryear {%
Lundberg%
, Erion%
\BCBL {}\ \BBA {} Lee%
}{%
Lundberg%
, Erion%
\BCBL {}\ \BBA {} Lee%
}{%
{\protect \APACyear {2018}}%
}]{%
lundberg2018consistent}
\APACinsertmetastar {%
lundberg2018consistent}%
\begin{APACrefauthors}%
Lundberg, S\BPBI M.%
, Erion, G\BPBI G.%
\BCBL {}\ \BBA {} Lee, S\BHBI I.%
\end{APACrefauthors}%
\unskip\
\newblock
\APACrefYearMonthDay{2018}{}{}.
\newblock
{\BBOQ}\APACrefatitle {Consistent individualized feature attribution for tree
  ensembles} {Consistent individualized feature attribution for tree
  ensembles}.{\BBCQ}
\newblock
\APACjournalVolNumPages{arXiv preprint arXiv:1802.03888}{}{}{}.
\PrintBackRefs{\CurrentBib}

\bibitem [\protect \citeauthoryear {%
Lundberg%
, Nair%
\BCBL {}\ \protect \BOthers {.}}{%
Lundberg%
, Nair%
\BCBL {}\ \protect \BOthers {.}}{%
{\protect \APACyear {2018}}%
}]{%
lundberg2018explainable}
\APACinsertmetastar {%
lundberg2018explainable}%
\begin{APACrefauthors}%
Lundberg, S\BPBI M.%
, Nair, B.%
, Vavilala, M\BPBI S.%
, Horibe, M.%
, Eisses, M\BPBI J.%
, Adams, T.%
\BDBL {}others%
\end{APACrefauthors}%
\unskip\
\newblock
\APACrefYearMonthDay{2018}{}{}.
\newblock
{\BBOQ}\APACrefatitle {Explainable machine-learning predictions for the
  prevention of hypoxaemia during surgery} {Explainable machine-learning
  predictions for the prevention of hypoxaemia during surgery}.{\BBCQ}
\newblock
\APACjournalVolNumPages{Nature biomedical engineering}{2}{10}{749--760}.
\PrintBackRefs{\CurrentBib}

\bibitem [\protect \citeauthoryear {%
Mannering%
, Bhat%
, Shankar%
\BCBL {}\ \BBA {} Abdel-Aty%
}{%
Mannering%
\ \protect \BOthers {.}}{%
{\protect \APACyear {2020}}%
}]{%
mannering2020big}
\APACinsertmetastar {%
mannering2020big}%
\begin{APACrefauthors}%
Mannering, F.%
, Bhat, C\BPBI R.%
, Shankar, V.%
\BCBL {}\ \BBA {} Abdel-Aty, M.%
\end{APACrefauthors}%
\unskip\
\newblock
\APACrefYearMonthDay{2020}{}{}.
\newblock
{\BBOQ}\APACrefatitle {Big data, traditional data and the tradeoffs between
  prediction and causality in highway-safety analysis} {Big data, traditional
  data and the tradeoffs between prediction and causality in highway-safety
  analysis}.{\BBCQ}
\newblock
\APACjournalVolNumPages{Analytic methods in accident research}{25}{}{100113}.
\PrintBackRefs{\CurrentBib}

\bibitem [\protect \citeauthoryear {%
McDonald%
, Lee%
, Schwarz%
\BCBL {}\ \BBA {} Brown%
}{%
McDonald%
\ \protect \BOthers {.}}{%
{\protect \APACyear {2014}}%
}]{%
mcdonald2014steering}
\APACinsertmetastar {%
mcdonald2014steering}%
\begin{APACrefauthors}%
McDonald, A\BPBI D.%
, Lee, J\BPBI D.%
, Schwarz, C.%
\BCBL {}\ \BBA {} Brown, T\BPBI L.%
\end{APACrefauthors}%
\unskip\
\newblock
\APACrefYearMonthDay{2014}{}{}.
\newblock
{\BBOQ}\APACrefatitle {Steering in a random forest: Ensemble learning for
  detecting drowsiness-related lane departures} {Steering in a random forest:
  Ensemble learning for detecting drowsiness-related lane departures}.{\BBCQ}
\newblock
\APACjournalVolNumPages{Human factors}{56}{5}{986--998}.
\PrintBackRefs{\CurrentBib}

\bibitem [\protect \citeauthoryear {%
Munkhdalai%
, Wang%
, Park%
\BCBL {}\ \BBA {} Ryu%
}{%
Munkhdalai%
\ \protect \BOthers {.}}{%
{\protect \APACyear {2019}}%
}]{%
munkhdalai2019advanced}
\APACinsertmetastar {%
munkhdalai2019advanced}%
\begin{APACrefauthors}%
Munkhdalai, L.%
, Wang, L.%
, Park, H\BPBI W.%
\BCBL {}\ \BBA {} Ryu, K\BPBI H.%
\end{APACrefauthors}%
\unskip\
\newblock
\APACrefYearMonthDay{2019}{}{}.
\newblock
{\BBOQ}\APACrefatitle {Advanced Neural Network Approach, Its Explanation with
  LIME for Credit Scoring Application} {Advanced neural network approach, its
  explanation with lime for credit scoring application}.{\BBCQ}
\newblock
\BIn{} \APACrefbtitle {Asian Conference on Intelligent Information and Database
  Systems} {Asian conference on intelligent information and database systems}\
  (\BPGS\ 407--419).
\PrintBackRefs{\CurrentBib}

\bibitem [\protect \citeauthoryear {%
Murdoch%
, Singh%
, Kumbier%
, Abbasi-Asl%
\BCBL {}\ \BBA {} Yu%
}{%
Murdoch%
\ \protect \BOthers {.}}{%
{\protect \APACyear {2019}}%
}]{%
murdoch2019definitions}
\APACinsertmetastar {%
murdoch2019definitions}%
\begin{APACrefauthors}%
Murdoch, W\BPBI J.%
, Singh, C.%
, Kumbier, K.%
, Abbasi-Asl, R.%
\BCBL {}\ \BBA {} Yu, B.%
\end{APACrefauthors}%
\unskip\
\newblock
\APACrefYearMonthDay{2019}{}{}.
\newblock
{\BBOQ}\APACrefatitle {Definitions, methods, and applications in interpretable
  machine learning} {Definitions, methods, and applications in interpretable
  machine learning}.{\BBCQ}
\newblock
\APACjournalVolNumPages{Proceedings of the National Academy of
  Sciences}{116}{44}{22071--22080}.
\PrintBackRefs{\CurrentBib}

\bibitem [\protect \citeauthoryear {%
Nagabushanam%
, George%
\BCBL {}\ \BBA {} Radha%
}{%
Nagabushanam%
\ \protect \BOthers {.}}{%
{\protect \APACyear {2019}}%
}]{%
nagabushanam2019eeg}
\APACinsertmetastar {%
nagabushanam2019eeg}%
\begin{APACrefauthors}%
Nagabushanam, P.%
, George, S\BPBI T.%
\BCBL {}\ \BBA {} Radha, S.%
\end{APACrefauthors}%
\unskip\
\newblock
\APACrefYearMonthDay{2019}{}{}.
\newblock
{\BBOQ}\APACrefatitle {EEG signal classification using LSTM and improved neural
  network algorithms} {Eeg signal classification using lstm and improved neural
  network algorithms}.{\BBCQ}
\newblock
\APACjournalVolNumPages{Soft Computing}{}{}{1--23}.
\PrintBackRefs{\CurrentBib}

\bibitem [\protect \citeauthoryear {%
Ribeiro%
, Singh%
\BCBL {}\ \BBA {} Guestrin%
}{%
Ribeiro%
\ \protect \BOthers {.}}{%
{\protect \APACyear {2016}}%
}]{%
ribeiro2016should}
\APACinsertmetastar {%
ribeiro2016should}%
\begin{APACrefauthors}%
Ribeiro, M\BPBI T.%
, Singh, S.%
\BCBL {}\ \BBA {} Guestrin, C.%
\end{APACrefauthors}%
\unskip\
\newblock
\APACrefYearMonthDay{2016}{}{}.
\newblock
{\BBOQ}\APACrefatitle {" Why should i trust you?" Explaining the predictions of
  any classifier} {" why should i trust you?" explaining the predictions of any
  classifier}.{\BBCQ}
\newblock
\BIn{} \APACrefbtitle {Proceedings of the 22nd ACM SIGKDD international
  conference on knowledge discovery and data mining} {Proceedings of the 22nd
  acm sigkdd international conference on knowledge discovery and data mining}\
  (\BPGS\ 1135--1144).
\PrintBackRefs{\CurrentBib}

\bibitem [\protect \citeauthoryear {%
SAE%
}{%
SAE%
}{%
{\protect \APACyear {2018}}%
}]{%
sae2018taxonomy}
\APACinsertmetastar {%
sae2018taxonomy}%
\begin{APACrefauthors}%
SAE.%
\end{APACrefauthors}%
\unskip\
\newblock
\APACrefYear{2018}.
\newblock
\APACrefbtitle {Taxonomy and Definitions for Terms Related to Driving
  Automation Systems for On-Road Motor Vehicles} {Taxonomy and definitions for
  terms related to driving automation systems for on-road motor vehicles}.
\newblock
\APACaddressPublisher{}{{SAE} International in United States, J3016--201806}.
\PrintBackRefs{\CurrentBib}

\bibitem [\protect \citeauthoryear {%
Sayed%
\ \BBA {} Eskandarian%
}{%
Sayed%
\ \BBA {} Eskandarian%
}{%
{\protect \APACyear {2001}}%
}]{%
sayed2001unobtrusive}
\APACinsertmetastar {%
sayed2001unobtrusive}%
\begin{APACrefauthors}%
Sayed, R.%
\BCBT {}\ \BBA {} Eskandarian, A.%
\end{APACrefauthors}%
\unskip\
\newblock
\APACrefYearMonthDay{2001}{}{}.
\newblock
{\BBOQ}\APACrefatitle {Unobtrusive drowsiness detection by neural network
  learning of driver steering} {Unobtrusive drowsiness detection by neural
  network learning of driver steering}.{\BBCQ}
\newblock
\APACjournalVolNumPages{Proceedings of the Institution of Mechanical Engineers,
  Part D: Journal of Automobile Engineering}{215}{9}{969--975}.
\PrintBackRefs{\CurrentBib}

\bibitem [\protect \citeauthoryear {%
Shapley%
, Kuhn%
\BCBL {}\ \BBA {} Tucker%
}{%
Shapley%
\ \protect \BOthers {.}}{%
{\protect \APACyear {1953}}%
}]{%
shapley1953contributions}
\APACinsertmetastar {%
shapley1953contributions}%
\begin{APACrefauthors}%
Shapley, L\BPBI S.%
, Kuhn, H.%
\BCBL {}\ \BBA {} Tucker, A.%
\end{APACrefauthors}%
\unskip\
\newblock
\APACrefYearMonthDay{1953}{}{}.
\newblock
{\BBOQ}\APACrefatitle {Contributions to the Theory of Games} {Contributions to
  the theory of games}.{\BBCQ}
\newblock
\APACjournalVolNumPages{Annals of mathematics studies}{28}{2}{307--317}.
\PrintBackRefs{\CurrentBib}

\bibitem [\protect \citeauthoryear {%
Sikander%
\ \BBA {} Anwar%
}{%
Sikander%
\ \BBA {} Anwar%
}{%
{\protect \APACyear {2018}}%
}]{%
sikander2018driver}
\APACinsertmetastar {%
sikander2018driver}%
\begin{APACrefauthors}%
Sikander, G.%
\BCBT {}\ \BBA {} Anwar, S.%
\end{APACrefauthors}%
\unskip\
\newblock
\APACrefYearMonthDay{2018}{}{}.
\newblock
{\BBOQ}\APACrefatitle {Driver fatigue detection systems: A review} {Driver
  fatigue detection systems: A review}.{\BBCQ}
\newblock
\APACjournalVolNumPages{IEEE Transactions on Intelligent Transportation
  Systems}{20}{6}{2339--2352}.
\PrintBackRefs{\CurrentBib}

\bibitem [\protect \citeauthoryear {%
Sun%
, Yu%
, Berilla%
, Liu%
\BCBL {}\ \BBA {} Wu%
}{%
Sun%
\ \protect \BOthers {.}}{%
{\protect \APACyear {2011}}%
}]{%
sun2011vehicle}
\APACinsertmetastar {%
sun2011vehicle}%
\begin{APACrefauthors}%
Sun, Y.%
, Yu, X.%
, Berilla, J.%
, Liu, Z.%
\BCBL {}\ \BBA {} Wu, G.%
\end{APACrefauthors}%
\unskip\
\newblock
\APACrefYearMonthDay{2011}{}{}.
\newblock
{\BBOQ}\APACrefatitle {An in-vehicle physiological signal monitoring system for
  driver fatigue detection} {An in-vehicle physiological signal monitoring
  system for driver fatigue detection}.{\BBCQ}
\newblock
\BIn{} \APACrefbtitle {3rd International Conference on Road Safety and
  SimulationPurdue UniversityTransportation Research Board.} {3rd international
  conference on road safety and simulationpurdue universitytransportation
  research board.}
\PrintBackRefs{\CurrentBib}

\bibitem [\protect \citeauthoryear {%
{\"U}nal%
, de Waard%
, Epstude%
\BCBL {}\ \BBA {} Steg%
}{%
{\"U}nal%
\ \protect \BOthers {.}}{%
{\protect \APACyear {2013}}%
}]{%
unal2013driving}
\APACinsertmetastar {%
unal2013driving}%
\begin{APACrefauthors}%
{\"U}nal, A\BPBI B.%
, de Waard, D.%
, Epstude, K.%
\BCBL {}\ \BBA {} Steg, L.%
\end{APACrefauthors}%
\unskip\
\newblock
\APACrefYearMonthDay{2013}{}{}.
\newblock
{\BBOQ}\APACrefatitle {Driving with music: Effects on arousal and performance}
  {Driving with music: Effects on arousal and performance}.{\BBCQ}
\newblock
\APACjournalVolNumPages{Transportation research part F: traffic psychology and
  behaviour}{21}{}{52--65}.
\PrintBackRefs{\CurrentBib}

\bibitem [\protect \citeauthoryear {%
Vogelpohl%
, K{\"u}hn%
, Hummel%
\BCBL {}\ \BBA {} Vollrath%
}{%
Vogelpohl%
\ \protect \BOthers {.}}{%
{\protect \APACyear {2019}}%
}]{%
vogelpohl2019asleep}
\APACinsertmetastar {%
vogelpohl2019asleep}%
\begin{APACrefauthors}%
Vogelpohl, T.%
, K{\"u}hn, M.%
, Hummel, T.%
\BCBL {}\ \BBA {} Vollrath, M.%
\end{APACrefauthors}%
\unskip\
\newblock
\APACrefYearMonthDay{2019}{}{}.
\newblock
{\BBOQ}\APACrefatitle {Asleep at the automated wheel—Sleepiness and fatigue
  during highly automated driving} {Asleep at the automated wheel—sleepiness
  and fatigue during highly automated driving}.{\BBCQ}
\newblock
\APACjournalVolNumPages{Accident Analysis \& Prevention}{126}{}{70--84}.
\PrintBackRefs{\CurrentBib}

\bibitem [\protect \citeauthoryear {%
Watta%
, Lakshmanan%
\BCBL {}\ \BBA {} Hou%
}{%
Watta%
\ \protect \BOthers {.}}{%
{\protect \APACyear {2007}}%
}]{%
watta2007nonparametric}
\APACinsertmetastar {%
watta2007nonparametric}%
\begin{APACrefauthors}%
Watta, P.%
, Lakshmanan, S.%
\BCBL {}\ \BBA {} Hou, Y.%
\end{APACrefauthors}%
\unskip\
\newblock
\APACrefYearMonthDay{2007}{}{}.
\newblock
{\BBOQ}\APACrefatitle {Nonparametric approaches for estimating driver pose}
  {Nonparametric approaches for estimating driver pose}.{\BBCQ}
\newblock
\APACjournalVolNumPages{IEEE transactions on vehicular
  technology}{56}{4}{2028--2041}.
\PrintBackRefs{\CurrentBib}

\bibitem [\protect \citeauthoryear {%
Zhang%
, Wang%
\BCBL {}\ \BBA {} Fu%
}{%
Zhang%
\ \protect \BOthers {.}}{%
{\protect \APACyear {2013}}%
}]{%
zhang2013automated}
\APACinsertmetastar {%
zhang2013automated}%
\begin{APACrefauthors}%
Zhang, C.%
, Wang, H.%
\BCBL {}\ \BBA {} Fu, R.%
\end{APACrefauthors}%
\unskip\
\newblock
\APACrefYearMonthDay{2013}{}{}.
\newblock
{\BBOQ}\APACrefatitle {Automated detection of driver fatigue based on entropy
  and complexity measures} {Automated detection of driver fatigue based on
  entropy and complexity measures}.{\BBCQ}
\newblock
\APACjournalVolNumPages{IEEE Transactions on Intelligent Transportation
  Systems}{15}{1}{168--177}.
\PrintBackRefs{\CurrentBib}

\bibitem [\protect \citeauthoryear {%
Zhou%
\ \protect \BOthers {.}}{%
Zhou%
\ \protect \BOthers {.}}{%
{\protect \APACyear {2020}}%
}]{%
zhou2020driver}
\APACinsertmetastar {%
zhou2020driver}%
\begin{APACrefauthors}%
Zhou, F.%
, Alsaid, A.%
, Blommer, M.%
, Curry, R.%
, Swaminathan, R.%
, Kochhar, D.%
\BDBL {}Lei, B.%
\end{APACrefauthors}%
\unskip\
\newblock
\APACrefYearMonthDay{2020}{}{}.
\newblock
{\BBOQ}\APACrefatitle {Driver Fatigue Transition Prediction in Highly Automated
  Driving Using Physiological Features} {Driver fatigue transition prediction
  in highly automated driving using physiological features}.{\BBCQ}
\newblock
\APACjournalVolNumPages{Expert Systems with Applications}{}{}{113204}.
\PrintBackRefs{\CurrentBib}

\bibitem [\protect \citeauthoryear {%
Zhou%
, Yang%
\BCBL {}\ \BBA {} Zhang%
}{%
Zhou%
\ \protect \BOthers {.}}{%
{\protect \APACyear {2019}}%
}]{%
zhou2019takeover}
\APACinsertmetastar {%
zhou2019takeover}%
\begin{APACrefauthors}%
Zhou, F.%
, Yang, X\BPBI J.%
\BCBL {}\ \BBA {} Zhang, X.%
\end{APACrefauthors}%
\unskip\
\newblock
\APACrefYearMonthDay{2019}{}{}.
\newblock
{\BBOQ}\APACrefatitle {Takeover Transition in Autonomous Vehicles: A {Y}ou
  {T}ube Study} {Takeover transition in autonomous vehicles: A {Y}ou {T}ube
  study}.{\BBCQ}
\newblock
\APACjournalVolNumPages{International Journal of Human--Computer
  Interaction}{}{}{1--12}.
\PrintBackRefs{\CurrentBib}

\end{thebibliography}

\end{document}